\documentclass[sn-mathphys-num]{sn-jnl}

\usepackage{graphicx}%
\usepackage{multirow}%
\usepackage{amsmath,amssymb,amsfonts}%
\usepackage{amsthm}%
\usepackage{mathrsfs}%
\usepackage[title]{appendix}%
\usepackage{xcolor}%
\usepackage{textcomp}%
\usepackage{manyfoot}%
\usepackage{booktabs}%
\usepackage{algorithmicx}%
\usepackage{algpseudocode}%
\usepackage{listings}%

\usepackage[linesnumbered,ruled,vlined]{algorithm2e}
\usepackage{subfig}

\theoremstyle{thmstyleone}%
\newtheorem{theorem}{Theorem}
\theoremstyle{thmstyletwo}%

\theoremstyle{thmstylethree}%

\begin{document}

\title[Article Title]{Multi-task Federated Learning with Encoder-Decoder Structure: Enabling Collaborative Learning Across Different Tasks}


\author[1]{\fnm{Jingxuan} \sur{Zhou}}\email{zhoujingxuan@nudt.edu.cn}

\author*[1]{\fnm{Weidong} \sur{Bao}}\email{wdbao@nudt.edu.cn}

\author[1]{\fnm{Ji} \sur{Wang}}\email{wangji@nudt.edu.cn}

\author[1]{\fnm{Dayu} \sur{Zhang}}\email{zhangdayu18@nudt.edu.cn}

\author[1]{\fnm{Xiongtao} \sur{Zhang}}\email{zhangxiongtao14@nudt.edu.cn}

\author[1]{\fnm{Yaohong} \sur{Zhang}}

\affil[1]{\orgdiv{Laboratory for Big Data and Decision}, \orgname{National University of Defense Technology}, \orgaddress{\city{Changsha}, \postcode{410073}, \state{Hunan}, \country{China}}}




\abstract{Federated learning has been extensively studied and applied due to its ability to ensure data security in distributed environments while building better models. However, clients participating in federated learning still face limitations, as clients with different structures or tasks cannot participate in learning together. In view of this, constructing a federated learning framework that allows collaboration between clients with different model structures and performing different tasks, enabling them to share valuable knowledge to enhance model efficiency, holds significant practical implications for the widespread application of federated learning. To achieve this goal, we propose a multi-task federated learning with encoder-decoder structure (M-Fed). Specifically, given the widespread adoption of the encoder-decoder architecture in current models, we leverage this structure to share intra-task knowledge through traditional federated learning methods and extract general knowledge from the encoder to achieve cross-task knowledge sharing. The training process is similar to traditional federated learning, and we incorporate local decoder and global decoder information into the loss function. The local decoder iteratively updates and gradually approaches the global decoder until sufficient cross-task knowledge sharing is achieved. Our method is lightweight and modular, demonstrating innovation compared to previous research. It enables clients performing different tasks to share general knowledge while maintaining the efficiency of traditional federated learning systems. We conducted experiments on two widely used benchmark datasets to verify the feasibility of M-Fed and compared it with traditional methods. The experimental results demonstrate the effectiveness of M-Fed in multi-task federated learning.}




\maketitle

\section{Introduction}\label{sec1}

In recent decades, deep neural networks have been widely applied in numerous real-world scenarios due to their remarkable success in solving complex problems across different domains. For instance, as stated in~\cite{Zhou2016}, tasks such as recognition, segmentation, and image parsing constitute the core components of medical image analysis.
In the field of autonomous driving, to navigate safely in the surrounding environment, it is imperative to detect and analyze all objects in the scene, accurately locate these objects, and estimate their distances and trajectories~\cite{Vandenhende2022}.
The continuous surge of other innovative applications is driving an exponential growth in the number of machine learning models specifically designed to meet diverse demands. Nowadays, it has become the norm to construct specific deep learning models for various types of tasks.

Currently, although large language models capable of executing diverse tasks have emerged, due to the limited computational capabilities of most devices, a more prevalent practice in practical applications is still to customize exclusive models for specific types of tasks.
However, building deep learning models for specific tasks requires extensive training datasets and computational resources, posing considerable challenges for individual users.
To enable individual users to build efficient models even with limited training data and computational resources, Federated Learning (FL) is proposed as a solution, which can also effectively protect the privacy of user data~\cite{Konecny2015}.

Despite its numerous advantages, FL also exhibits certain limitations in practical applications, primarily manifested in two aspects: data scarcity and model constraints:
\begin{enumerate}
\item{\emph{data scarcity}: For tasks with high specificity and a limited number of available clients, FL faces difficulties in acquiring sufficient high-quality data from a restricted number of clients during implementation. This leads to constraints on improving model performance. Therefore, traditional FL methods still encounter challenges in constructing high-performance models for specific tasks with scarce data.}
\item{\emph{model constraints}: Traditional FL methods require all clients to perform the same task and possess the same model structure. This implies that when clients participate in the FL system, they must collaborate with a group of clients performing the same task and sharing the same model structure. Although knowledge distillation~\cite{Hinton2015} techniques provide clients with the flexibility to operate using models with different structures, the application of FL in diverse tasks still faces significant challenges. The requirement for task homogeneity in FL limits its capacity to accommodate a larger number of clients.}
\end{enumerate}

Our objective is to design a framework that addresses the limitations of traditional FL, enabling clients with diverse tasks and model structures to participate in FL collectively, thereby naturally mitigating the issue of data scarcity.
We observe that the current model construction process tends to be streamlined, commonly adopting an encoder-decoder architecture.
In this architecture, the encoder is responsible for converting variable-length input sequences into fixed-length vectors, while the decoder performs the opposite operation, generating variable-length sequences from fixed-length vectors.
The encoder, as a fundamental component for feature extraction, demonstrates adaptability to various downstream tasks; by contrast, the decoder is customized for specific downstream tasks to better meet their professional requirements.
By combining a unified encoder with different decoders, multitask adaptation can be achieved under similar network architectures.

We propose a method named \textbf{M}ulti-task \textbf{F}ederated Learning with \textbf{E}ncoder-\textbf{D}ecoder Structure (M-Fed). This method aims to break through the inherent limitations of data scarcity and model constraints in traditional FL, enabling clients with different tasks and diverse model structures to collaborate and learn from each other, thereby effectively alleviating the issue of scarce task data. As a lightweight and scalable technology, M-Fed allows clients to participate in FL without being restricted by model structures and task types, simply by making local adjustments to their original models.
Furthermore, to enhance training efficiency, we introduce a variable threshold mechanism into the actual training process. This mechanism can effectively mitigate the negative impact of the global model on local models during the initial stage of training.

The differences between previous studies and this research are shown in Figure~\ref{Concept}. Previous studies primarily focused on collaborative learning among clients with identical models, while the M-Fed proposed in this research faces two unique challenges in collaborative learning: 1) constructing a framework to enable collaborative learning among multi-task clients; 2) effectively transferring knowledge between heterogeneous models. To address these challenges, we have designed a multi-task collaborative learning framework aimed at minimizing the differences between heterogeneous models and achieving knowledge transfer without compromising the execution of local tasks. The main contributions and innovations of this paper can be summarized as follows:

\begin{figure}[htbp]
\centering
\includegraphics[width=\textwidth]{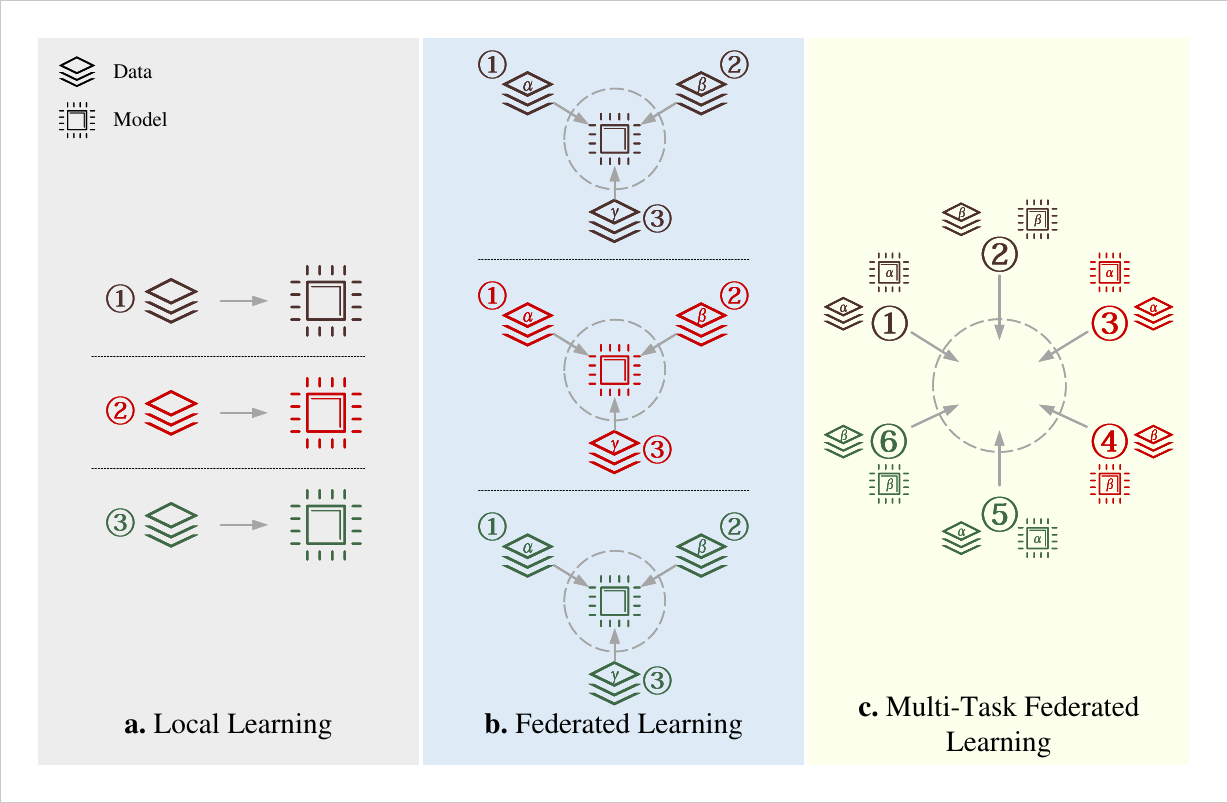}
\caption{Comparison of multi-task FL with other typical methods. Different colors represent different types of tasks. \textbf{a.} represents the case where local data is used for training models and there is no interaction between clients, with the quality of the models entirely reliant on the quality of the local data. \textbf{b.} demonstrates the restriction of traditional FL that prohibits cross-learning between client models with diverse structures executing various tasks. \textbf{c.} indicates the M-Fed method can transfer general knowledge to improve performance across different tasks, even if they have different model architectures on multiple clients.}
\label{Concept}
\end{figure}

\begin{itemize}
\item{We propose the M-Fed method, a FL technique tailored for multi-task scenarios. This method enables cooperative learning across heterogeneous clients for different tasks, while retaining the privacy-preserving advantages of traditional FL approaches. Clients only need to make local adjustments to their original models to participate in M-Fed, without altering their original tasks.}
\item{The M-Fed method is based on the widely used encoder-decoder architecture. Under this architecture, we can extract common knowledge from models with different tasks and structures and share it with all participating clients, thereby enhancing the performance of all models. Furthermore, we provide a proof of convergence for the M-Fed method.}
\item{To validate the effectiveness of the M-Fed method, we conducted experiments on four different tasks using the PASCAL Context and MS COCO datasets. The experimental results demonstrate that the M-Fed method can significantly improve the accuracy of models for various tasks.}
\end{itemize}

The rest of this paper is organized as follows: Section~\ref{sec2} briefly overviews the related literature on multi-task learning and traditional FL; Section~\ref{sec3} introduces the specific framework of the M-Fed method in detail; Section~\ref{sec4}, we conduct a thorough theoretical analysis of the proposed M-Fed algorithm; Section~\ref{sec5} describes the experimental setup and results, and compares the proposed method with standard methods; Section~\ref{sec6}, we discuss the limitations of the method proposed in this study; finally, Section~\ref{sec7} summarizes the paper and suggests possible directions for future research.

\section{Related Work}\label{sec2}
This section begins with an overview of previous research on multi-task learning, aiming to sort out and clarify the differing viewpoints of scholars. Subsequently, we provide an outline of the operational framework of traditional FL and its development trends.

\subsection{Multi-Task Learning}
Regarding the concept of "tasks" in multi-task learning, there are different viewpoints in the academic community. Smith et al.~\cite{Smith2017} argue that the goal of multi-task learning is to build models that can handle multiple related tasks simultaneously. Under this framework, even if the model structures are the same, they are considered as separate tasks as long as there are significant differences in data distribution~\cite{Argyriou2006, Chen2011, Goncalves2016, Evgeniou2004, Jacob2008, Kim2009, Zhang2010}. This type of research treats different data distributions as multi-tasking, aiming to reveal the intrinsic connections between them. On the other hand, other scholars generally believe that "tasks" in multi-task learning are multiple learning tasks that are related but not entirely the same~\cite{Zhang2017}.

After clarifying the research scope of multi-task learning, scholars have turned their attention to its objectives. Some scholars believe that tasks in multi-task learning can mutually reinforce each other. Caruana~\cite{Caruana1997} points out that multi-task and single-task learning use the same input but have different outputs. By capturing and leveraging correlations between tasks, multi-task learning can extract general knowledge, thereby enhancing the effectiveness of task outputs. Xu et al.~\cite{Xu2018} further emphasize the crucial role of inter-task correlations in improving overall performance. At the same time, some scholars view multi-task learning as a strategy to enhance specific tasks. Research by Ran et al.~\cite{Ran2023} shows that introducing additional related tasks can improve the performance of the main task.

Other scholars consider mitigating data sparsity as the objective of multi-task learning~\cite{Zhang2021}. They argue that the amount of labeled data provided by users is insufficient for training high-precision models, while multi-task learning can aggregate labeled data from all related tasks to improve the accuracy of each task model. This method is also seen as an effective way to reuse common knowledge and reduce the amount of labeled data required for model training.

In the study of multi-task learning, the relationship between tasks is also a significant aspect, which can be broadly classified into two categories. The first category assumes that the correlations between tasks are known apriori~\cite{Chen2011, Argyriou2006, Evgeniou2004, Kim2009}. The second category believes that the interrelationships between tasks are unknown and need to be derived from the data~\cite{Zhang2010, Jacob2008, Goncalves2016, Zhou2021}. This research primarily focuses on the former category because, in a single collaborative learning system, it is feasible to arrange related tasks based on their specific attributes.

Multi-task learning provides a valuable approach for model training by transferring common knowledge across different tasks. This characteristic makes it particularly suitable for FL, where models are trained on data across multiple devices and tasks. Despite this potential, existing FL methods still have limitations in adapting to distributed scenarios and multi-tasking.
This research emphasizes another dimension of multi-task learning—the learning mechanism across different models, particularly distributed multi-task learning. This method focuses on solving the problem of how to conduct effective multi-task FL when data and tasks are distributed across multiple clients. Compared to training a single global model, the distributed multi-task learning approach can promote the sharing of general knowledge between tasks, thereby enhancing model performance.

\subsection{Federated Learning}
FL, as an efficient method, enables data utilization and model construction among multiple participants while ensuring user privacy protection and data security. Traditional FL typically assumes that all clients use the same model as the central server and perform the same tasks. Based on this assumption, researchers have conducted extensive studies in various aspects such as aggregation algorithms~\cite{BrendanMcMahan2016, Li2020a, Karimireddy2020}, privacy security~\cite{Song2023, Wang2018, Huang2020}, trust~\cite{Majeed2019}, and incentive mechanisms~\cite{Kang2019}, aiming to improve the efficiency of FL and expand its application scope.

However, as research progresses, researchers have gradually recognized the issues posed by the common assumption of models in FL, namely, that participants need to use the same model and perform the same tasks. This undoubtedly imposes numerous constraints on the large-scale application of FL. Consequently, many scholars have begun to ponder the interplay between FL and localization. Fallah et al.~\cite{Fallah2020} proposed learning an appropriate initialization model to enable participants to quickly construct their local models. Liu et al.~\cite{Liu2022a} investigated the impact of differential privacy on cross-silo FL and explored the implications of multi-task processing, laying the foundation for private cross-silo FL and indicating future research directions in this field. He et al.~\cite{He2022} introduced a new multi-task federated training framework capable of operating on molecular graphs with incomplete labels and lacking a centralized server. In FedRep~\cite{Collins2021}, the authors suggested utilizing all available data from clients to learn global representations, which not only enhances the performance of the global model but also allows clients to perform multiple local updates to enhance their generalization ability to local data. Ditto~\cite{Li2021} proposed a multi-task framework that addresses the competing constraints of accuracy, fairness, and robustness in FL.

These studies have made significant progress in promoting the practical application of FL. However, the aforementioned multi-task FL methods are still rooted in the traditional FL architecture, where the system assumes that participants share the same model. Although these studies consider factors such as multi-task processing and personalization, there are some fundamental differences between such issues and the multi-task problems we are currently investigating. Our core objective is to break away from the traditional FL architecture and delve into the exploration of multi-task FL from the perspective of task model constraints.
\section{Our Proposed Method}\label{sec3}
\subsection{Problem Description}
The fundamental objective of traditional FL is to integrate models trained by multiple clients using their local data, thereby forming a unified global model. Specifically:
\begin{equation}
\label{FL}
\begin{split}
&\min\limits_{}{\sum\limits_{i \in N}{L_i\left( x_{i},y_{i};w_{i} \right)}} \\ 
&G = A(F(w_{1}),F(w_{2}),\cdots , F(w_{N})), 
\end{split}
\end{equation}
where ${{x}_{i}}$ and ${{y}_{i}}$ respectively represent the local data and labels of client $i$. ${{w}_{i}}$ denotes the set of parameters for client $i$. The loss function $L_i(\cdot )$ is the objective that client $i$ needs to minimize. After completing a specified number of iterations, each client uploads its parameters to the server. The function $A(\cdot)$ aggregates the local models $F({{w}_{i}}) $ from all clients to form a global model $G$.

Let us consider the scenario of multi-task collaborative learning, where we assume the existence of a set of clients $N$ and a set of tasks $K$, satisfying the condition $K \leq N$.
Under this setting, each client $i \in N$ is responsible for executing a specific task, and different clients may perform the same task.
Each client $i$ holds a local dataset consisting of data and label pairs $x_{i}, y_{i}$.
Specifically, clients performing the same task adopt the same model structure, while clients with different tasks may adopt either the same or different model structures.
The client determines the loss function $L_{k}(x_{i},y_{i};w_{n})$ based on the task. When client $i$ performs task $k$, the model trained using local data can be represented as $F_{k}(w_{i})$.
Traditional FL aggregation methods require all clients to adopt the same model structure, which leads to the inability to directly aggregate models from different clients in multi-task scenarios. Specifically, when $j\neq i$, $F_{i}(\cdot)$ and $F_{j}(\cdot)$ cannot be aggregated. Addressing the challenge of cooperative learning between such heterogeneous models is formidable.

\subsection{Framework of M-Fed}
\begin{figure}[htbp]
\centering
\includegraphics[width=\textwidth]{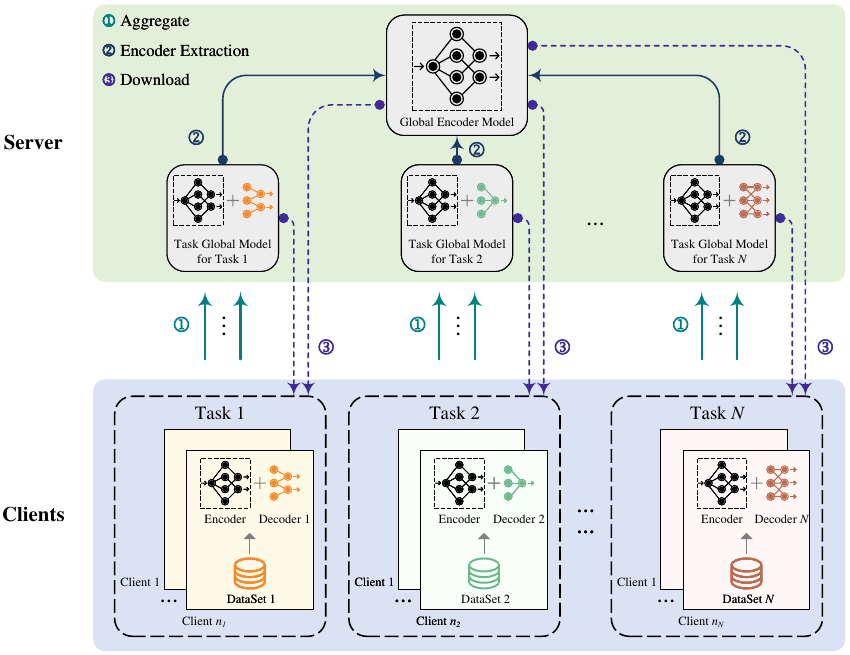}
\caption{The overall framework of our proposed method. The \textbf{client} set consists of clients performing different tasks. These clients primarily utilize local data and combine it with the task global model and global encoder model obtained from the server to train their local models. In this process, all clients adopt a unified encoder, while the decoder is customized according to each client's specific task. After a specific number of local training rounds, the clients upload the trained encoders and decoders to the server.
Upon receiving the models uploaded by the clients, the \textbf{server} aggregates the models performing the same task to form a task global model. Subsequently, the server extracts and aggregates the encoder model components from all task global models to form a global encoder model. Finally, the server sends the global encoder model to all clients and sends the task global model to the clients corresponding to each specific task.}\label{framework}

\end{figure}

Our method, illustrated in Figure~\ref{framework}, addresses the challenge of how clients with diverse tasks and model structures can engage in FL. To facilitate knowledge sharing, it is necessary to align models across different tasks. Observations reveal that the current model construction process commonly adopts an encoder-decoder architecture, where the encoder processes general knowledge and the decoder handles specific downstream tasks. Therefore, by unifying the encoder structure of participating client models while allowing the decoder to remain task-specific, we can achieve alignment of general knowledge across different tasks, thereby promoting collaborative learning.

Our designed M-Fed framework is as follows: Clients participating in multi-task FL need to adjust their models, with the encoder part being uniformly specified by the server and the decoder component being customizable to meet the requirements of specific tasks. Clients performing the same task use the same decoder. The learning process initially follows the traditional FL paradigm, where clients train their models based on local data and then upload the trained models to the server. After receiving all client models, the server aggregates those performing the same task to form a task-global model, enabling collaborative learning within the same task. This process is relatively straightforward due to the consistent model structure for executing the same task. Subsequently, for models performing different tasks, since we have unified the encoder model structure, we can aggregate the encoder parts to form a global encoder model, achieving cross-task collaborative learning. After receiving the global encoder model and task-global models from the server, clients need to integrate these inter-task and intra-task knowledge into their local models and then upload them to the server. Through repeated iterations, the effectiveness of all models is enhanced. A detailed elaboration of this process will be provided in the next section.

Furthermore, once the server disseminates the global encoder model to the clients, the clients will perform a predetermined number of iterations using local data. During these iterations, the global encoder model remains unchanged since it is not updated, which somewhat hinders the convergence of the client models to the convergence point. To mitigate the negative impact of the global encoder model on the client models, we have also designed an adaptive parameter that can effectively manage the influence of the global encoder model on the client models. When significant changes occur in the parameters of the client model, we believe that the global encoder model is too far from the convergence point, and the influence of the global encoder model on the client model should be reduced. Once the parameters of the client model stabilize, we believe that the global encoder is close to the convergence point, and at this time, the influence of the global encoder model on the client model can be increased. This effectively reduces the negative impact of the global encoder model on the client models, making the training process of multi-task FL smoother and more efficient.

\subsection{M-Fed Algorithm}

This section will elaborate on the specific implementation process of the M-Fed algorithm. As described in the previous section, the overall framework of M-Fed indicates that after receiving the task global model and the global encoder model sent by the server, clients need to utilize these models to update their local models, facilitating knowledge fusion within tasks and knowledge transfer across tasks. For knowledge fusion within tasks, we can draw lessons from traditional FL methods, namely, directly replacing the local model with the task global model. Since the local model and the task global model are structurally identical, this step is easy to implement. Through this approach, clients can acquire knowledge from other clients within the same task. Regarding knowledge transfer across tasks, although the structure of the global encoder model is consistent with the encoder structure of the local model, considering the differences between tasks, directly replacing the local encoder model with the global encoder model is not advisable, as it may lead to a decline in the performance of some tasks. Therefore, we propose to achieve knowledge transfer across tasks by reducing the distance between the encoder part of the local model and the global encoder model. To realize multi-task FL, we suggest adding the following objectives based on traditional joint learning goals:
\begin{equation}
\label{obj}
\min{\sum\limits_{i \in N}\left\| {w_{i}^{*} - g} \right\|_{2}},
\end{equation}
where $w_{i}^{*}$ represents the encoder component in client $i$'s model, and $g$ denotes the global encoder model. By minimizing the distance between the encoder components of all client models and the global encoder model, the transfer of general knowledge can be effectively facilitated. Next, we will elaborate on the implementation of this process.

Firstly, we elaborate on the generation process of the global encoder model. Upon receiving all models uploaded by clients, the server aggregates models that perform the same task. The aggregation process can be represented as:

\begin{equation}
\begin{split}
& F_{k}(w) = A\left( {F_{k}\left( w_{1} \right),F_{k}\left( w_{2} \right),\cdots,F_{k}\left( w_{N_{k}} \right)} \right), \\
&s.t.\quad  \left\{\begin{array}{lc}
    k \in K\\
    \end{array}\right.
\end{split}
\end{equation}
where, $N_k$ denotes the number of clients performing task $k$. Through the aggregation algorithm, all models performing the same task are aggregated, ultimately forming a set of $|K|$ global models for different tasks on the server side, denoted as $\left\{ {F_{1},F_{2},\cdots,F_{|K|}} \right\}$. Next, we extract the encoder component from each global task model to construct the global encoder model. The specific process is as follows:
\begin{equation}
g = A\left( {F_{1}^{*}\left( w^{*} \right),F_{2}^{*}\left( w^{*} \right),\cdots,F_{|K|}^{*}\left( w^{*} \right)} \right),
\end{equation}
where, $F_{k}^{*}\left( w^{*} \right)$ represents the encoder component of the model for task $k$.
Given that the encoder components of all tasks share the same structure, we can derive the global encoder model $g$ using the specified aggregation algorithm.
In related tasks, the shallow layers of the model exhibit certain commonalities~\cite{Kokkinos2017}. Despite the differences between tasks, relevant clients can still achieve knowledge sharing across different tasks by aggregating shallow parameters.

Next, we examine the behavior of clients. Similar to traditional FL, clients need to perform a certain number of local training rounds using their local data. Assuming that each client $i$ holds its unique local dataset and performs task $k$, the objective of its local training can be expressed as:
\begin{equation}
\begin{split}
&\min {\sum\limits_{i \in N_{k}}{L_{i}\left( {x_{i},y_{i};w_{i}} \right)}} \\ 
&L_{i}\left( {x_{i},y_{i};w_{i}} \right) = \frac{S_{i}}{S}{\underset{w_{i}}{\arg\min}f_{i}}\left( {{x_{i},y}_{i};w_{i}} \right),
\end{split}
\end{equation}
where $f_i(\cdot)$ represents the loss function of client $i$, $S$ denotes the total amount of data across all clients, while $S_i$ specifically refers to the amount of data held by client $i$. The core goal of clients is to explore a set of parameters that can minimize the loss value on their local datasets. This objective is aligned with the pursuit of traditional deep learning methods.

How does the client achieve knowledge sharing? Whenever the server distributes the global model for a task, the client replaces its local model with this global model, thereby achieving knowledge sharing within the task. It is important to note that directly replacing the encoder part of the local model with the global encoder model is not advisable. To acquire knowledge across tasks, we introduce a regularization term during the local training process, with the aim of making the parameters of the encoder part of the local model as close as possible to those of the global encoder model. The specific process is as follows:
\begin{equation}
{\min\limits_{w_{i}}{h\left( {{x_{i},y}_{i};w_{i}} \right)}} = L_{i}\left( {x_{i},y_{i};w_{i}} \right) + \lambda\left\| {w_{i}^{*} - g} \right\|_{2},
\end{equation}
where $w_{i}^{*}$ represents the encoder component of client $i$'s parameter $w_{i}$, which shares the same structure as $g$ and is therefore suitable for norm calculation.
By introducing a regularization term into the original loss function, the client is able to effectively acquire general knowledge from different tasks without replacing the encoder part of the local model with the global encoder model. This strategy ensures that the performance of the original task is not affected, while also enabling the extraction of knowledge from other tasks.
The hyperparameter $\lambda$ serves as a mediator between the local model and the global encoder model. When $\lambda$ is decreased to near 0, the client learns knowledge solely from other clients performing the same task. Conversely, as $\lambda$ increases, the influence of the global objective becomes stronger. To optimize the training process, we can adjust the value of $\lambda$. In the early stages of training, due to significant fluctuations in local model parameters, excessive proximity between the local model and the global encoder model may reduce training efficiency. However, as training progresses, the changes in local model parameters gradually stabilize. At this point, by increasing the value of $\lambda$, we can encourage the local encoder to align more closely with the global encoder model, which facilitates the acquisition of general knowledge and ultimately enhances training efficiency.

\begin{algorithm}[htbp]
\SetAlgoLined 
\caption{M-Fed}
\label{alg}
\KwIn{$K$: Task set; $T$: Total rounds of aggregation; $N$: Client set; $m$: Local training rounds; $N_k$: The set of clients that perform task $k$, $k\in K$; $\lambda $: hyper-parameters; $\alpha $: Learning rate; $g$: Global encoder model; $w_k^0$: Initial parameter set;}
\KwOut{$w_k^{T-1}$, $k\in K$}
\For{$t=0,\cdots ,T-1$}{
    $w_k^t\to N_k $ \tcp*{The server allocates the parameter $w_k^t$ to the client that is executing task $k$.}
    $g\to N$ \tcp*{The server distributes the global encoder model $g$ to all clients.}
    \For{$n=1,\cdots ,N$ same}{
        $w_{n}^{t + 1} = w_{n}^{t} - \left( {\alpha\nabla F\left( w_{n}^{t} \right) + \lambda\left\| {w_{n}^{t, *} - g} \right\|_{2}} \right)$ \tcp*{Train the local model for $m$ rounds using local data.}
        Send $\mathrm{\Delta}w_{k,n}^{t} = w_{n}^{t + 1} - w_{n}^{t}$ to the server.
    }
    \For{$k=1,\cdots ,K$}{
    $\left. w_{k}^{t + 1}\leftarrow A\left( {w_{k}^{t};\mathrm{\Delta}w_{k,n}^{t}} \right) \right.$ \tcp*{Aggregate the parameters of task $k$ through the aggregation function $A$ to form the task global model.}
    }
  $\left. g\leftarrow A\left( {w_{1}^{t + 1,*},w_{2}^{t + 1,*},\cdots,w_{|K|}^{t + 1,*}} \right) \right.$ \tcp*{Aggregation function $A$ aggregates the encoder components of the task global models, thereby forming the global encoder model.}
}
return
\end{algorithm}

\section{Theoretical Analysis}\label{sec4}

The global encoder model $g$ does not rely on any task global model.
Therefore, $g$ exhibits the same global convergence rate as the local objectives. To analyze the convergence behavior of algorithm \ref{alg}, we provide a proof of its local convergence.

\begin{theorem}[Local Convergence of algorithm \ref{alg}, formal statement and proof in Theorem \ref{theorem}]\label{the1}
    For any $k \in K$, we assume that $F_k(\cdot)$ is a strongly convex and smooth function. Based on this assumption, there exists a constant $C < \infty$ such that the model $v_k$ for task $k$ can converge to its optimal model $v_k^*$ at a rate of $Cg(t)$.
\end{theorem}
Here, we redefine the symbols, for instance, using $*$ to denote the optimal model. This practice aims to conform to the conventions of proof and minimize ambiguity.
The complete statement of Theorem~\ref{the1} and its proof will be provided in detail in Appendix~\ref{app}. 

\section{Experimental Study}\label{sec5}
This section will provide a detailed introduction to the datasets we used and the corresponding experimental configurations. Subsequently, the effectiveness of our method is verified through experiments, and finally, the quantitative results of testing on different datasets are disclosed.

\subsection{Data Preparation and Setup}
\textit{Data sets}: This experiment primarily utilizes two major datasets: the MS COCO dataset~\cite{Lin2014} and the segmentation version of the PASCAL Context dataset~\cite{Mottaghi2014}. The MS COCO is a large-scale dataset widely used for object detection, segmentation, and captioning tasks, consisting of 328,000 images and 2.5 million labels. Specifically, in this study, we focus on four core tasks within the MS COCO dataset: instance segmentation, panoptic segmentation, keypoint detection, and stuff segmentation.
On the other hand, the segmentation version of the PASCAL Context dataset serves as a popular benchmark for dense prediction tasks. It includes 10,581 training images and 1,449 validation images, covering four main tasks: semantic segmentation, human part segmentation, saliency estimation, and semantic edge or boundary detection. Detailed information about these two datasets is presented in Table \ref{tab1}.

\begin{table*}[htbp]
    \centering
    \caption{Brief Description of Data Sets}
    \label{tab1}
    \resizebox{\textwidth}{!}{%
    \begin{tabular}{@{}l|ccc|cccccccc@{}}
    \toprule
    \textbf{Dataset} &
      \textbf{Train} &
      \textbf{Val} &
      \textbf{Test} &
      \textbf{Segm.} &
      \textbf{Key} &
      \textbf{Parts} &
      \textbf{Inst.} &
      \textbf{Sal.} &
      \textbf{Edges} &
      \textbf{Pano.} &
      \textbf{Stuff} \\ \midrule
    MS COCO &
      118,287 &
      5,000 &
      40,670 &
       &
      $\checkmark$ &
       &
      $\checkmark$ &
       &
       &
      $\checkmark$ &
      $\checkmark$ \\
    PASCAL-Context &
      10,581 &
      1,449 &
      - &
      $\checkmark$ &
       &
      $\checkmark$ &
       &
      $\checkmark$ &
      $\checkmark$ &
       &
       \\ \bottomrule
    \end{tabular}%
    }
\begin{tablenotes}
\footnotesize
\item Segm.: Semantic Segmentation, Key: Key-Point Detection, Parts: Human Parts Segmentation, Inst.: Instance Segmentation, Sal.: Saliency Estimation, Edges: Semantic Edges or Boundaries, Pano.: Panoptic Segmentation, Stuff: Stuff Segmentation.
\end{tablenotes}
\end{table*}

\textit{Model Setting}: The selection of the encoder in the model is determined by the server. In this experiment, we employed three different sizes of ResNet as encoders, specifically: ResNet-18, ResNet-50, and ResNet-101.
The four tasks performed on the PASCAL Context dataset are all based on pixel-level image classification, aiming to determine the category of each pixel in the image. For these tasks, we chose the DeepLabV3+ model and combined it with the ASPP module~\cite{Chen2018} as the decoder.
For the MS COCO dataset, different tasks utilized different decoders: the Mask R-CNN~\cite{He2017} was used as the decoder for the instance segmentation task; the Panoptic FPN~\cite{Kirillov2019a} served as the decoder for the panoptic segmentation task; the associative embedding method~\cite{Newell2017} was employed as the decoder for the key-point detection task; and DeepLabV3~\cite{Chen2017} was selected as the decoder for the Stuff Segmentation task.

\textit{Training Setup}: For the PASCAL-Context dataset, we configured 5 clients for each task type, ensuring equal and non-overlapping data distributions across clients. In all experiments, after completing 5 rounds of local training on the client devices, a model aggregation was performed, with a total of 20 aggregations. The variable threshold $\lambda$ exhibited an exponential growth trend based on the number of model aggregation rounds, gradually increasing its impact. Considering the relative simplicity of the PASCAL-Context dataset, we selected ResNet-18 as the encoder. Other components, such as the decoder, loss function, and optimizer, were adjusted according to the specific task requirements, with consistent parameters maintained for the same task.
For the MS COCO dataset, we also allocated 5 clients for each task, where each client had the same amount of training data without overlap. After each client completed one training round, a global aggregation was conducted, with a total of 12 aggregations. The variable threshold $\lambda$ also increased exponentially with the number of model aggregation rounds.
Given the larger number and more complex categorizations of images in the MS COCO dataset, we chose ResNet-50 and ResNet-101 as our encoders. The selection of decoders, loss functions, and related aspects was based on the specific requirements of the task at hand, with consistent parameters maintained for the same task.
Regarding the training rounds, each client in the PASCAL-Context dataset underwent a total of 100 training rounds, while each client in the MS COCO dataset only went through 12 rounds. This difference was attributed to the increased data volume and model parameters in the MS COCO dataset, leading to significantly different training times. However, this approach facilitated the validation of the effectiveness of our proposed algorithm from various perspectives.
The entire model was trained using PyTorch on 4 RTX 2080Ti GPUs.

\subsection{Evaluation Metric}
In our experiments, although the tasks vary in nature, they all involve pixel-level classification, leading to some common evaluation metrics. Specifically, the metric of \textit{odsF} (optimal Dataset scale F-measure)~\cite{Martin2004} is employed for the evaluation of edge detection tasks. The odsF metric is applied by setting a uniform threshold for all images, meaning that a fixed threshold is selected and consistently applied to all images with the aim of achieving the maximum F-measure across the entire dataset. This value is calculated using the formula $ \left( \frac{2 \cdot Precision \cdot Recall}{Precision + Recall} \right) $.

We adopt the metric of \textit{mIoU} (mean Intersection over Union)~\cite{Long2015} to evaluate task stuff segmentation, semantic segmentation, saliency estimation, and human part segmentation. The definition of \textit{mIoU} is as follows:
\begin{equation}
mIoU = \frac{1}{k}{\sum\limits_{i = 1}^{k}\frac{P\bigcap G}{P\bigcup G}},
\end{equation}
where $P$ is the prediction and $G$ is the ground truth.

Key-point and instance segmentation are evaluated using \textit{AP} (average precision)~\cite{Everingham2009}. The \textit{AP} can be defined as:
\begin{equation}
AP = {\int_{0}^{1}{PR(r)dr}} = {\sum_{n}{\left( R_{n} - R_{n - 1} \right)P_{n}}},
\end{equation}
where $P_{n}$ and $R_{n}$ represent the precision and recall metrics of the nth threshold, respectively. Precision $P$ can be defined as $P = T_{p} / (T_{p} + F_{p})$. $T_{p}$ stands for True Positives, which refers to the number of genuine instances that were correctly identified as positive, whereas $F_{p}$ stands for False Positives, which represents the number of incorrect positives predicted by the model. Recall $R$ can be defined as $R = T_{p} / (T_{p} + F_{n})$. $F_{n}$ refers to False Negatives, representing the number of actual positive instances missed by the model.

The Panoptic Segmentation task is evaluated using \textit{PQ} (panoptic quality)~\cite{Kirillov2019} as the evaluation metric. The \textit{PQ} can be defined as:
\begin{equation}
\label{PQ}
PQ = \underset{Detection~Quality(DQ)}{\underbrace{\frac{T_{p}}{T_{p} + \frac{1}{2}F_{p} + \frac{1}{2}F_{n}}}} \times \underset{Sementation~Quality(SQ)}{\underbrace{\frac{\sum_{(x,y) \in T_{p}}{IoU(x,y)}}{T_{p}}}},
\end{equation}
where, $T_{p}$, $F_{p}$, $ F_{n} $, etc. are consistent with the previous definitions. According to formula~\ref{PQ}, \textit{PQ} is composed of Detection Quality and Segmentation Quality components, divided into Instance-level and Pixel-level indicators. \textit{PQ} score is a uniform assessment of both indicators.

\subsection{Compared to Traditional Methods}
Our proposed framework for multi-task FL lacks comparable methods, thus we obtained local training results by training each individual task separately. For each independent task, we conducted traditional FL and obtained the results after FL for each task. To verify the effectiveness of our proposed method, we compared the performance of the M-Fed algorithm with that of the local training algorithm and the traditional FL method in our experiments. The experimental results on two different datasets are as follows:

\textit{Results on PASCAL-Context:} In the experimental results on the PASCAL Context dataset (see Table~\ref{PASCAL}, $\uparrow$ indicates that the larger the indicator, the better), our proposed M-Fed method demonstrates significant advantages compared to other comparative methods. Based on these observations, we draw the following conclusions:

\begin{table*}[htbp]
    \caption{Results on PASCAL-Context Dataset}
    \label{PASCAL}
    \resizebox{\textwidth}{!}{
    \begin{tabular}{l|l|cllllcllllcllllclllll}
    \toprule
    \textbf{Method} &
      \textbf{Encoder} &
      \multicolumn{5}{c}{\textbf{Segm.(\textit{mIoU}) $\uparrow $}} &
      \multicolumn{5}{c}{\textbf{Parts(\textit{mIoU}) $\uparrow $}} &
      \multicolumn{5}{c}{\textbf{Sal.(\textit{mIoU}) $\uparrow $}} &
      \multicolumn{5}{c}{\textbf{Edges(\textit{odsF}) $\uparrow $}} &
      \textbf{$ \Delta_{m} \% \uparrow $} \\ \midrule
    Local &
      \multirow{3}{*}{ResNet-18} &
      \multicolumn{5}{c}{42.07} &
      \multicolumn{5}{c}{50.74} &
      \multicolumn{5}{c}{58.50} &
      \multicolumn{5}{c}{\underline{62.60}} &
      + 0.00 \\
    Traditional FL &
       &
      \multicolumn{5}{c}{\underline{53.14}} &
      \multicolumn{5}{c}{\underline{55.76}} &
      \multicolumn{5}{c}{\underline{61.40}} &
      \multicolumn{5}{c}{\textbf{66.20}} &
      + 11.73 \\
    Ours &
       &
      \multicolumn{5}{c}{\textbf{53.98}} &
      \multicolumn{5}{c}{\textbf{55.93}} &
      \multicolumn{5}{c}{\textbf{61.56}} &
      \multicolumn{5}{c}{\textbf{66.20}} &
      \textbf{+ 12.38} \\ \bottomrule
    \end{tabular}}
    \end{table*}
\begin{itemize}
\item{Our proposed M-Fed method achieved the highest scores in all four tasks (\textit{mIoU} and \textit{odsF}). Specifically, we saw the most significant improvement in the Semantic Segmentation task, with a boost of $1.58\%$. The improvement in the Human Parts Segmentation task was $0.30\%$, and in the Saliency Estimation task was $0.26\%$. However, the improvement in the Semantic Edges or Boundaries task was not as pronounced.}
\item{The local results indicate that, in this task, the optimal evaluation metric values obtained by each client through model training using local data have been demonstrated. Compared to mere local training, the application of FL has significantly improved the evaluation scores. For instance, in the semantic segmentation task, benefiting from the parameter sharing achieved by the FL mechanism during the training process, which further facilitates indirect data sharing among tasks of the same type, the \textit{mIoU} value has increased by $26.31\%$. Furthermore, the M-Fed method takes a step further. It not only promotes parameter information sharing among models within the same task but also indirectly transfers data knowledge between different tasks, thereby effectively enhancing the overall performance of the model.}
\item{Our M-Fed method is a lightweight extension of traditional FL approaches, achieving functional enhancement without sacrificing the original performance. More specifically, experimental results demonstrate that semantic segmentation tasks exhibit robust generalization capabilities due to their extensive knowledge transferability. Both human part segmentation tasks and saliency estimation tasks have also achieved moderate performance improvements. However, although semantic edge or boundary tasks face significant challenges in absorbing transmitted information, they do not undermine the capabilities of traditional FL.}
\end{itemize}

\begin{figure}[htbp]
    \centering
    \subfloat[Semantic Segmentation]{
      \includegraphics[width=1.24in]{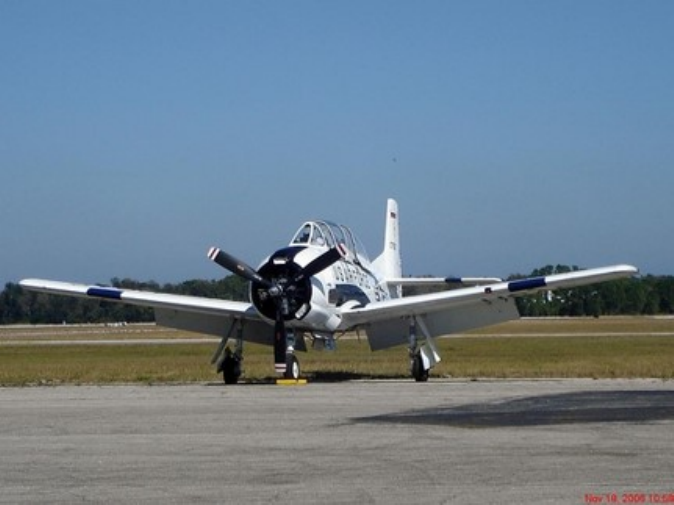}
      \includegraphics[width=1.24in]{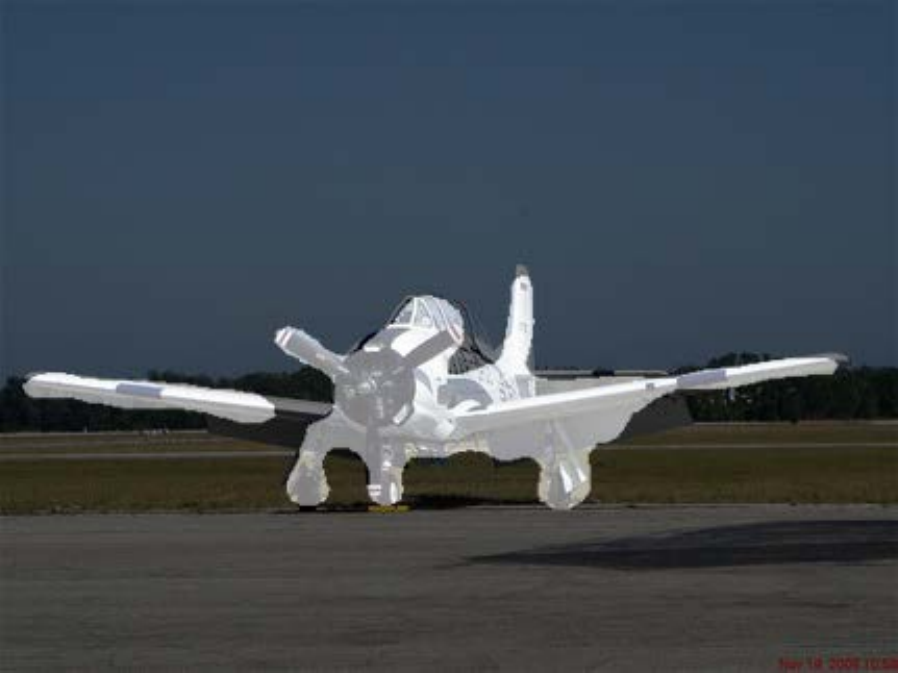}
      \includegraphics[width=1.24in]{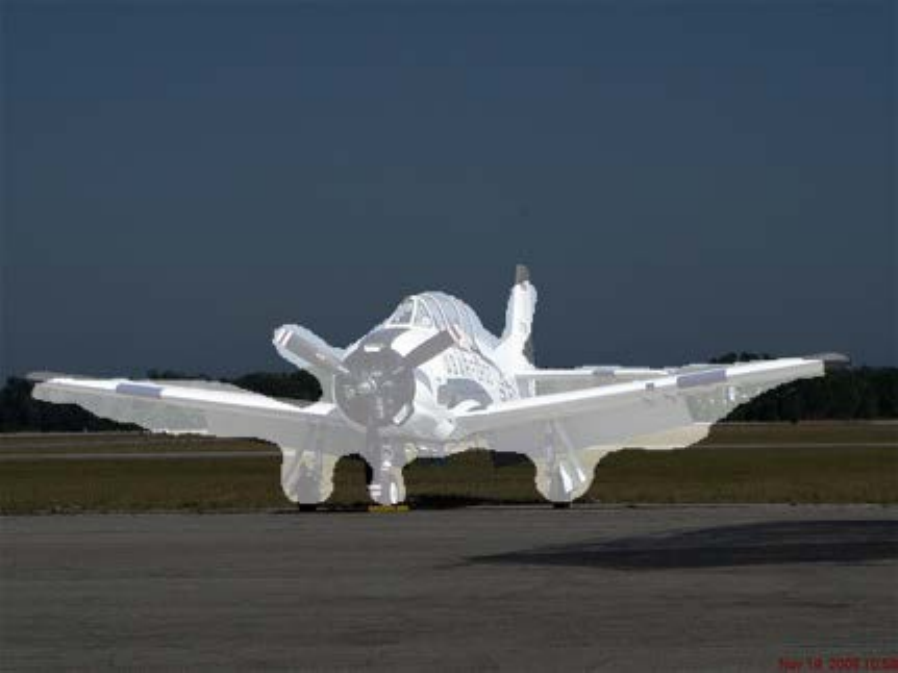}
      \includegraphics[width=1.24in]{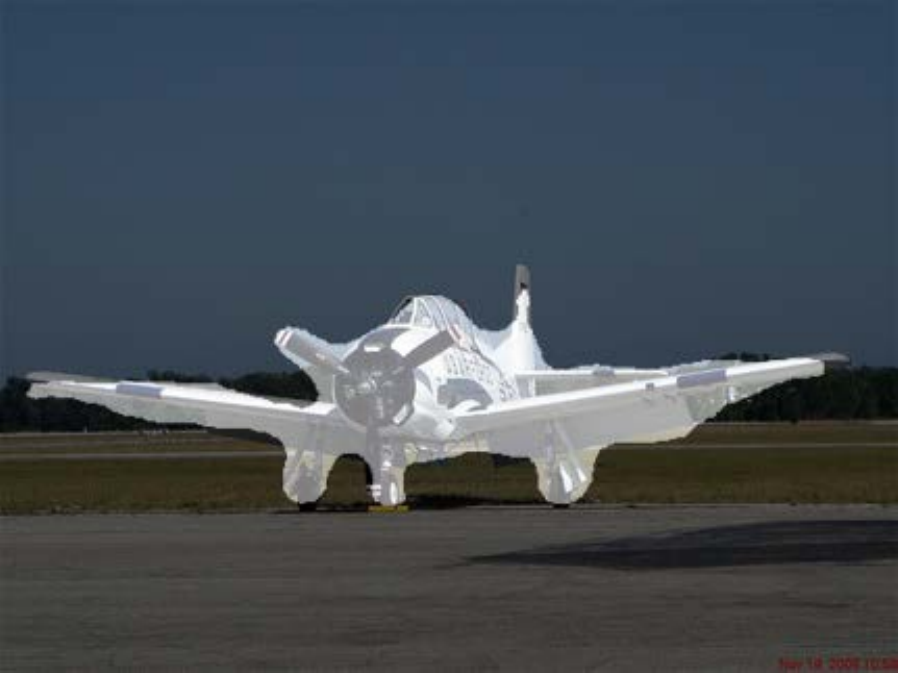}
    \label{segm}
    }
    \quad
    \subfloat[Human Parts Segmentation]{
      \includegraphics[width=1.24in]{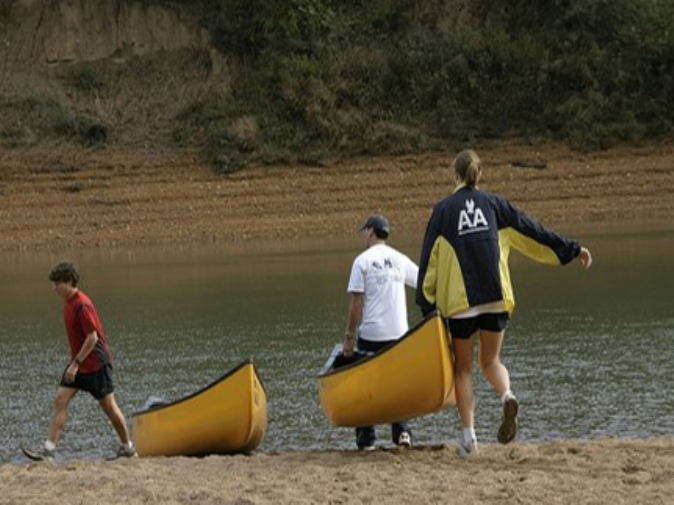}
      \includegraphics[width=1.24in]{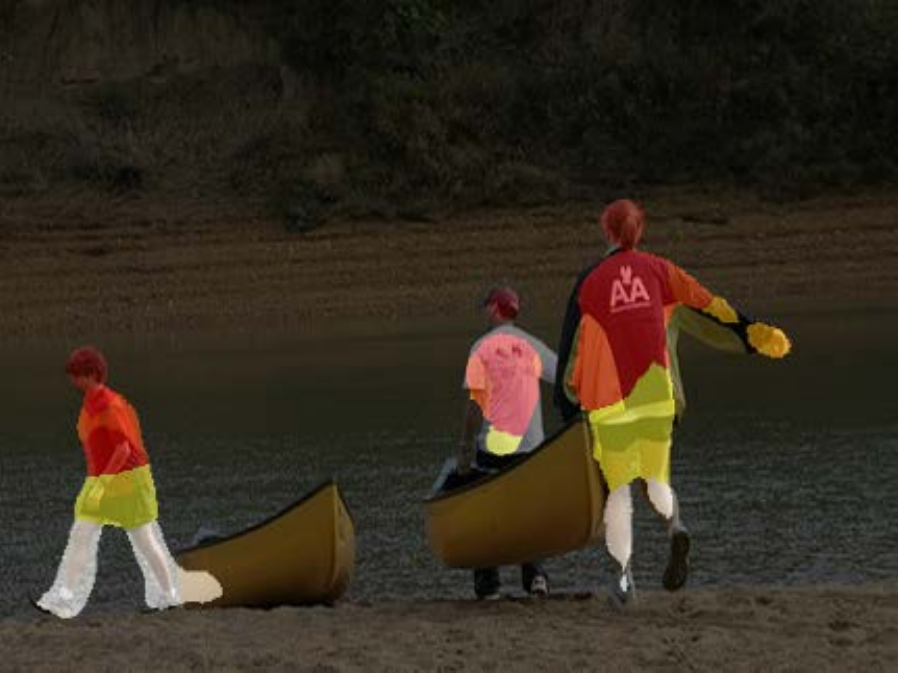}
      \includegraphics[width=1.24in]{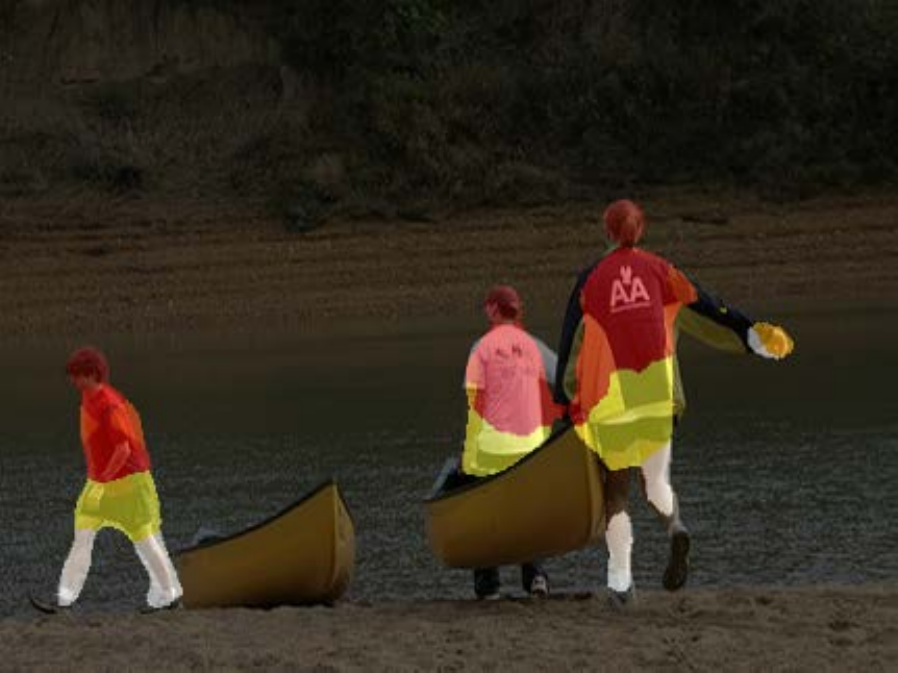}
      \includegraphics[width=1.24in]{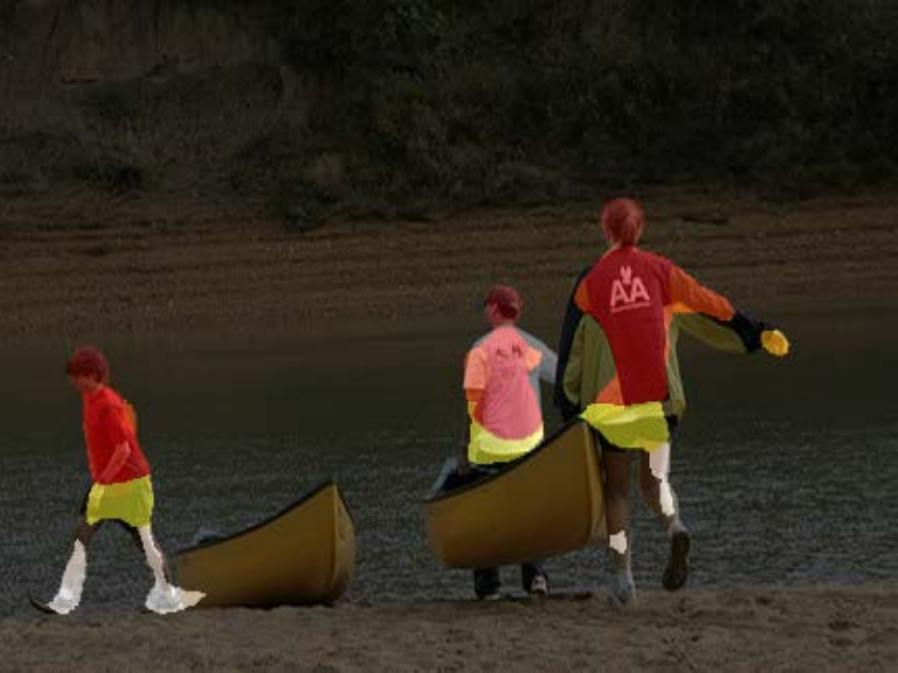}
    \label{human}
    }
    \quad
    \subfloat[Saliency Estimation]{
      \includegraphics[width=1.24in]{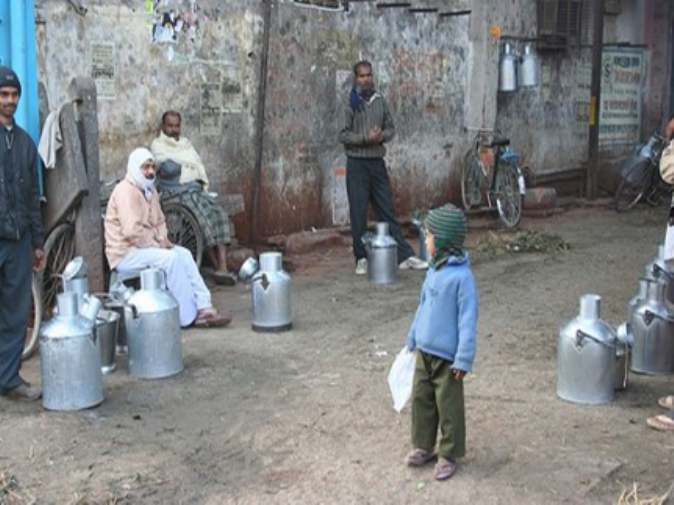}
      \includegraphics[width=1.24in]{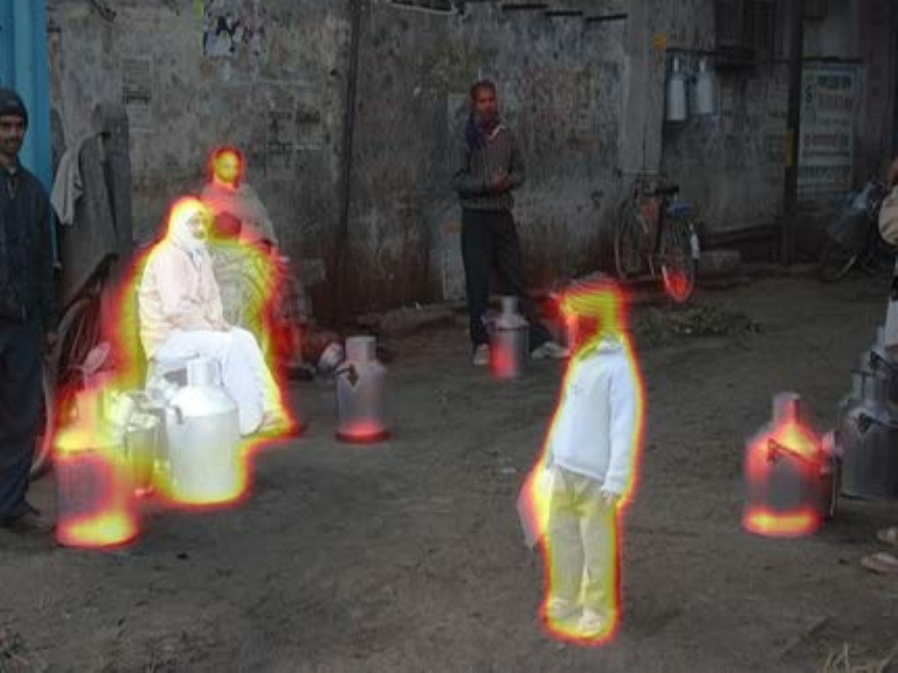}
      \includegraphics[width=1.24in]{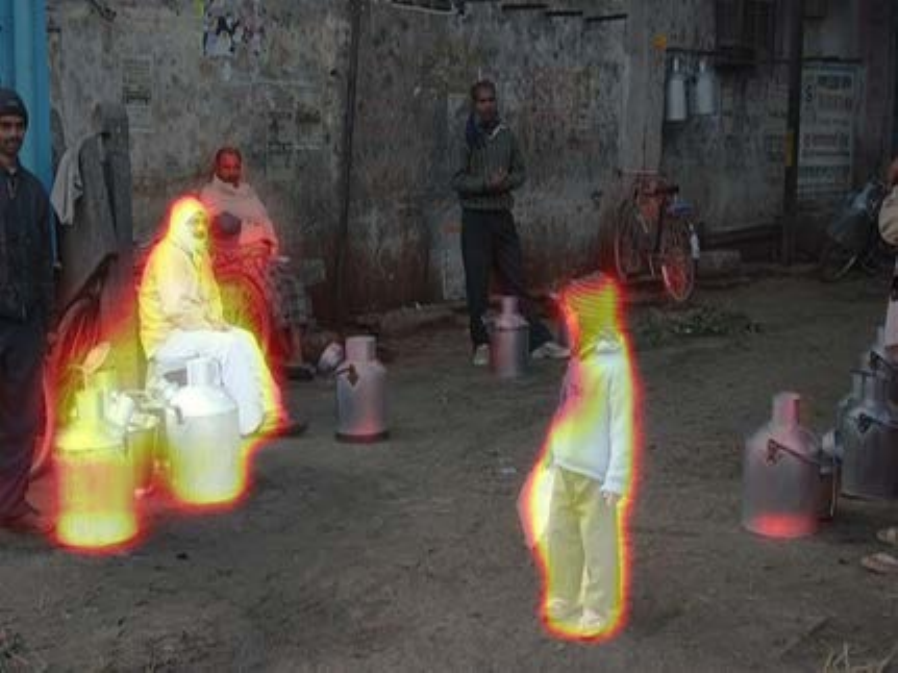}
      \includegraphics[width=1.24in]{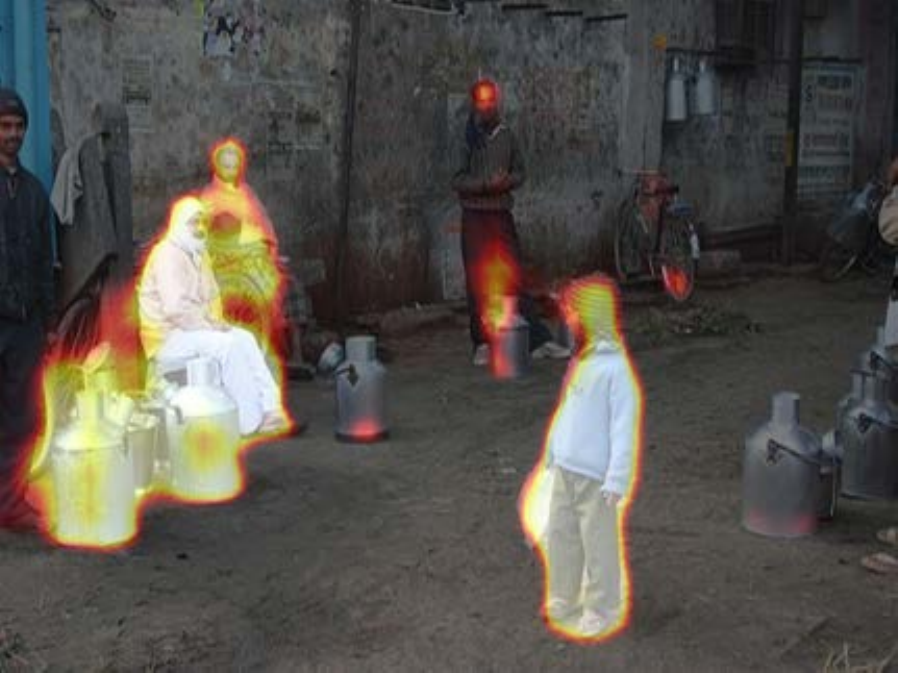}
    \label{sal}
    }
    \quad
    \subfloat[Semantic Edges or Boundaries]{
      \includegraphics[width=1.24in]{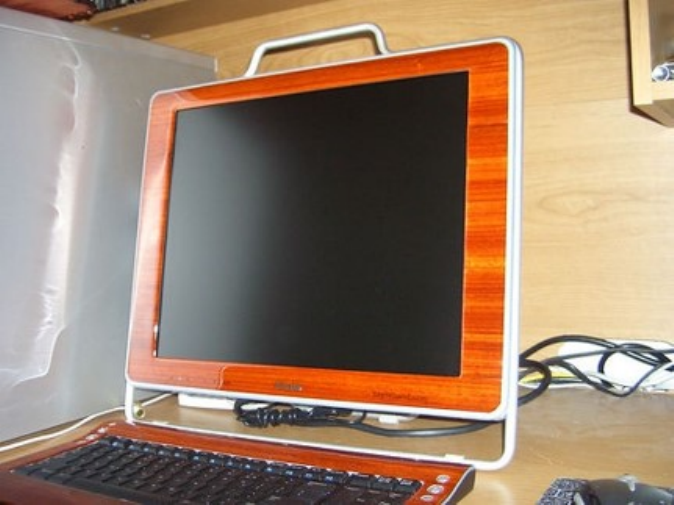}
      \includegraphics[width=1.24in]{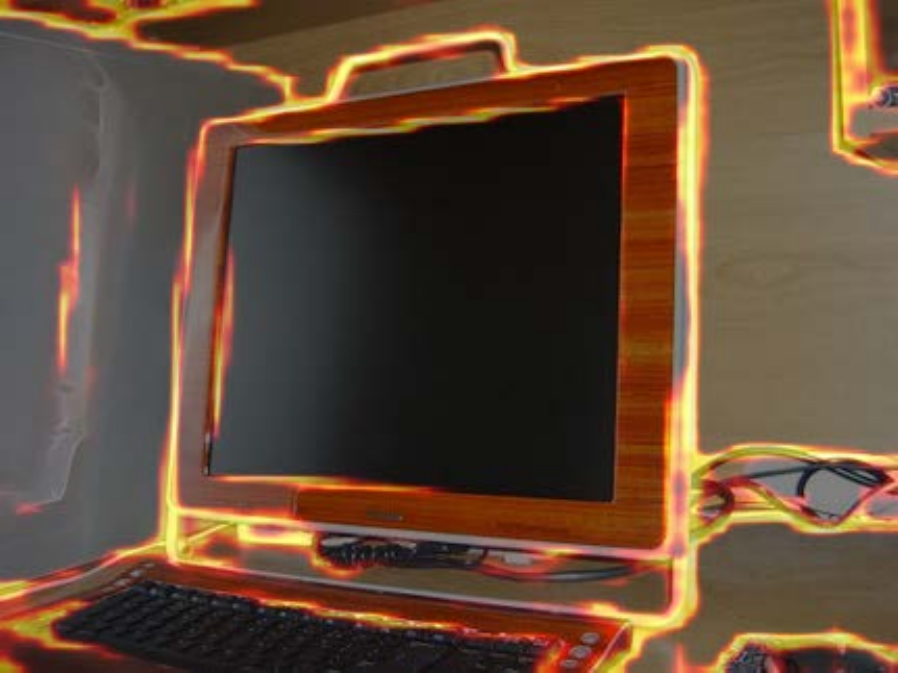}
      \includegraphics[width=1.24in]{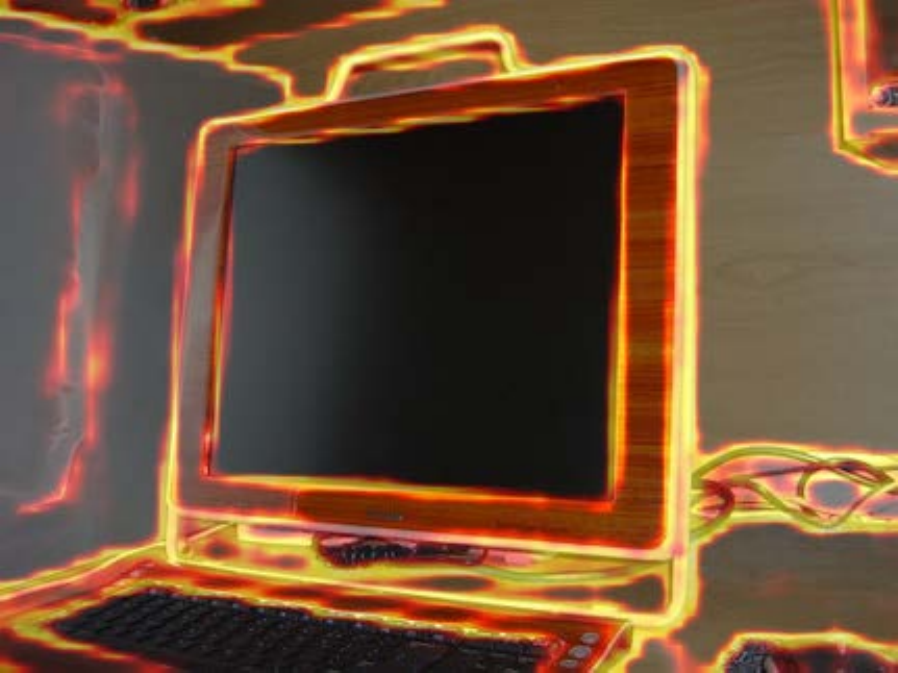}
      \includegraphics[width=1.24in]{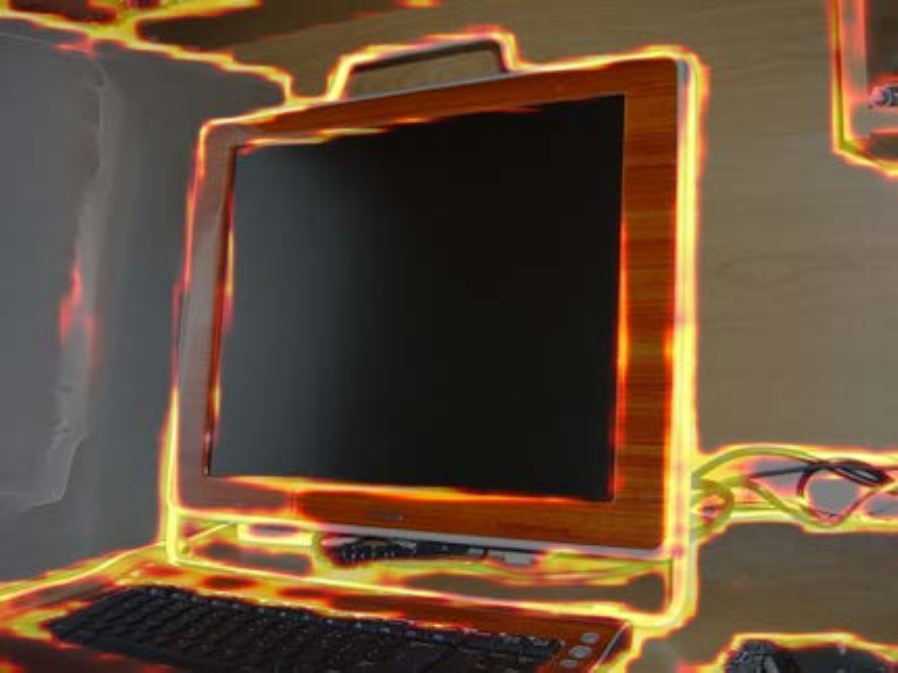}
    \label{edge}
    }
    \caption{Qualitative results on PASCAL-Context dataset. Each set of results shows the original image, local result, FL result, and our M-Fed result respectively from left to right.}
    \label{quantitative}
    \end{figure}

In the experiments conducted on the Pascal Context dataset, the M-Fed method exhibited differentiated results across different tasks. This phenomenon can potentially be attributed to the limited parameter capacity of ResNet-18 when used as an encoder, which hindered effective knowledge transfer between models.
To further delve into the effectiveness of the M-Fed method, we conducted additional quantitative analyses. Figure~\ref{quantitative} presents the quantitative experimental results obtained on the PASCAL Context dataset.

As evident from Figure~\ref{quantitative}, the M-Fed method demonstrates significant improvements in the quality of specific details across various tasks. Specifically, in the semantic segmentation task, our method exhibits notably superior details in object boundary delineation compared to the other two baseline methods. Similar advantages are observed in the human part segmentation task, particularly with more precise depictions of human torsos. For the saliency estimation task, our method is able to more effectively focus on key elements, filter out irrelevant information, and enhance the clarity of important contours. In the semantic edge or boundary detection task, irrelevant information on the left side of the computer is also significantly reduced.
In summary, although M-Fed does not show a significant increase in scores compared to traditional FL methods in Table~\ref{PASCAL}, it achieves the goal of multi-task FL. While focusing on their respective tasks, clients can also learn general knowledge from related tasks, thereby enhancing the overall effectiveness of all tasks.

\textit{Results on MS COCO:} The experimental results on the MS COCO dataset for M-Fed, traditional FL, and local training, using ResNet-50 and ResNet-101 as encoders, are presented in Table~\ref{COCO}. The results reveal the following conclusions:

\begin{table*}[htbp]
    \centering
    \caption{Results on MS COCO Dataset}
    \label{COCO}
    \resizebox{\textwidth}{!}{%
    \begin{tabular}{@{}l|c|cllllcllllcllllcllllc@{}}
    \toprule
    \textbf{Method} &
      \textbf{Encoder} &
      \multicolumn{5}{c}{\textbf{Inst.(\textit{AP}) $\uparrow $}} &
      \multicolumn{5}{c}{\textbf{Pano.(\textit{PQ}) $\uparrow $}} &
      \multicolumn{5}{c}{\textbf{Key(\textit{AP}) $\uparrow $}} &
      \multicolumn{5}{c}{\textbf{Stuff(\textit{mIoU}) $\uparrow $}} &
      \textbf{$ \Delta_{m} \% \uparrow $} \\ \midrule
    Local &
      \multirow{3}{*}{ResNet-50} &
      \multicolumn{5}{c}{18.97} &
      \multicolumn{5}{c}{\textbf{25.78}} &
      \multicolumn{5}{c}{26.61} &
      \multicolumn{5}{c}{5.37} &
      + 0.00 \\
    FL &
       &
      \multicolumn{5}{c}{\textbf{21.65}} &
      \multicolumn{5}{c}{23.67} &
      \multicolumn{5}{c}{\underline{31.17}} &
      \multicolumn{5}{c}{\underline{7.25}} &
      + 14.53 \\
    Ours &
       &
      \multicolumn{5}{c}{\underline{20.83}} &
      \multicolumn{5}{c}{\underline{24.21}} &
      \multicolumn{5}{c}{\textbf{32.69}} &
      \multicolumn{5}{c}{\textbf{7.40}} &
      \textbf{+ 16.09} \\ \midrule
    Local &
      \multirow{3}{*}{ResNet-101} &
      \multicolumn{5}{c}{20.38} &
      \multicolumn{5}{c}{\textbf{28.02}} &
      \multicolumn{5}{c}{33.40} &
      \multicolumn{5}{c}{15.45} &
      + 0.00 \\
    FL &
       &
      \multicolumn{5}{c}{\textbf{23.25}} &
      \multicolumn{5}{c}{25.92} &
      \multicolumn{5}{c}{\underline{39.89}} &
      \multicolumn{5}{c}{\underline{16.26}} &
      + 7.82 \\
    Ours &
       &
      \multicolumn{5}{c}{\underline{23.22}} &
      \multicolumn{5}{c}{\underline{25.96}} &
      \multicolumn{5}{c}{\textbf{40.33}} &
      \multicolumn{5}{c}{\textbf{16.62}} &
      \textbf{+ 8.73} \\ \bottomrule
    \end{tabular}%
    }
    \end{table*}

    \begin{itemize}
\item{Compared to local training methods, the M-Fed approach achieves significant overall performance improvements. Specifically, when using ResNet-50 as the encoder, the average score increases by 16.09\%, and when using ResNet-101, the average score rises by 8.73\%. Our method demonstrates exceptional performance on both encoders, achieving the highest scores in key-point detection and stuff segmentation tasks. It also outperforms traditional FL methods in panoptic segmentation tasks, albeit slightly inferior to them in instance segmentation tasks, yet still superior to local training methods.
Furthermore, in key-point detection tasks, M-Fed achieves a performance improvement of 4.88\% compared to traditional FL with ResNet-50 as the encoder and a 1.10\% improvement when compared to traditional FL with ResNet-101 as the encoder.
In stuff segmentation tasks, our method also exhibits advantages, achieving a performance improvement of 2.07\% compared to traditional FL when using ResNet-50 as the encoder and a 2.21\% improvement when using ResNet-101 as the encoder.
However, in instance segmentation tasks, the performance of M-Fed on both encoders decreases by 3.78\% and 0.13\% respectively compared to traditional FL, with relatively low decline rates. In panoptic segmentation tasks, the M-Fed method outperforms traditional FL on both encoders, albeit somewhat inferior to local training methods.}
\item{In the task of panoptic segmentation, regardless of whether the encoder is ResNet-50 or ResNet-101, the scores obtained by our M-Fed method and the classical FL are lower than those achieved by the local training method. Analysis suggests that this phenomenon can be attributed to the data distribution disparities among clients. Research indicates that uneven data distribution across clients may degrade the performance of FL~\cite{Zhao2018}. In the initial data allocation phase of our experiments, we did not consider the characteristics of data distribution. Instead, we simply scattered the data randomly and allocated an equal amount of data to each client, which resulted in uneven data distribution and subsequently affected the accuracy of FL. When comparing the M-Fed method with traditional FL, we observed that when using ResNet-50 as the encoder, our method achieved a performance improvement of 2.28\%, and when using ResNet-101 as the encoder, the performance also increased by 0.15\%. This indicates that although our method cannot completely eliminate the decline in system performance caused by data distribution issues, it still has a positive effect on improving the performance of traditional FL.}
\item{Our method outperforms local methods in the instance segmentation task but does not match the performance of the FL method. Further analysis reveals that the decline in the performance of the M-Fed method is closely related to the correlation between tasks. This method aims to facilitate clients performing different tasks to share common knowledge. However, when there are significant differences between tasks or a lack of shared knowledge, this method not only fails to improve client performance but may slightly weaken model performance. The instance segmentation task involves two core steps: firstly, predicting the bounding box of the target object, and secondly, conducting pixel-level classification within the predicted bounding box to determine its precise contour. Given that other tasks do not possess the relevant knowledge for detecting bounding boxes, our method demonstrates slightly inferior performance in some aspects compared to traditional FL methods. This experiment further reveals the limitations of the M-Fed method.}
\end{itemize}

\begin{figure*}[htbp]
    \centering
    \subfloat[Instance Segmentation]{
      \includegraphics[width=1.24in]{fig4-ins.pdf}
      \includegraphics[width=1.24in]{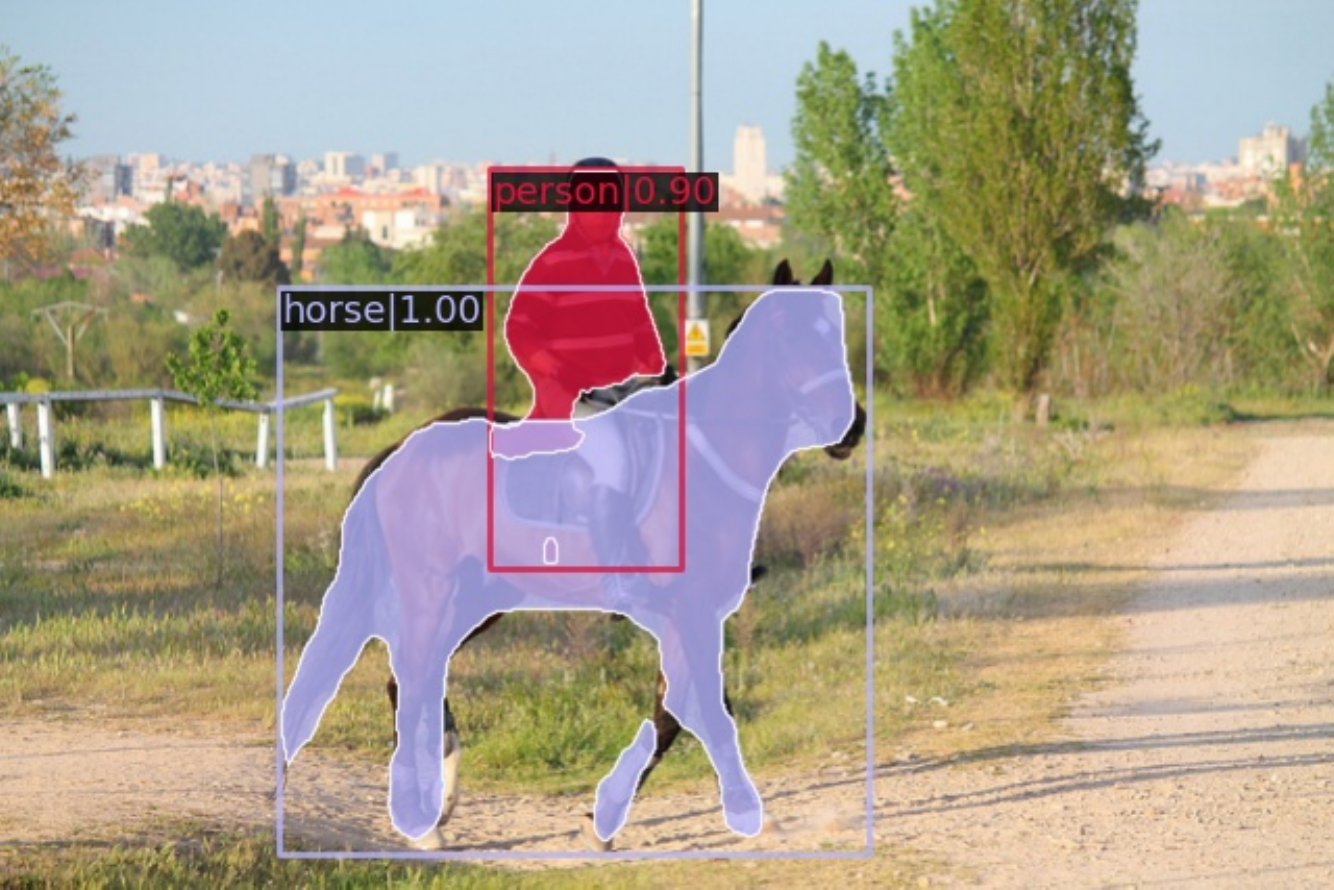}
      \includegraphics[width=1.24in]{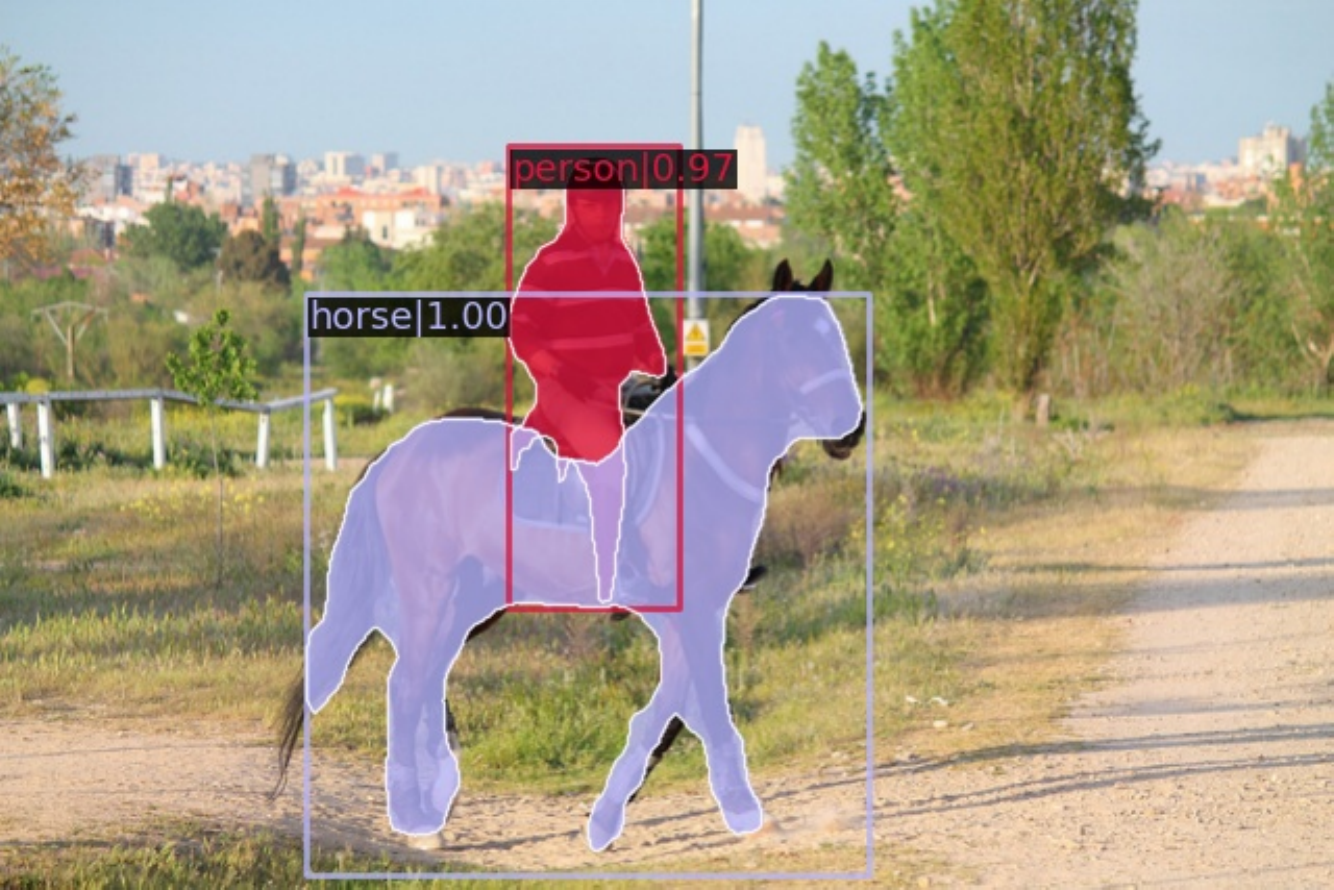}
      \includegraphics[width=1.24in]{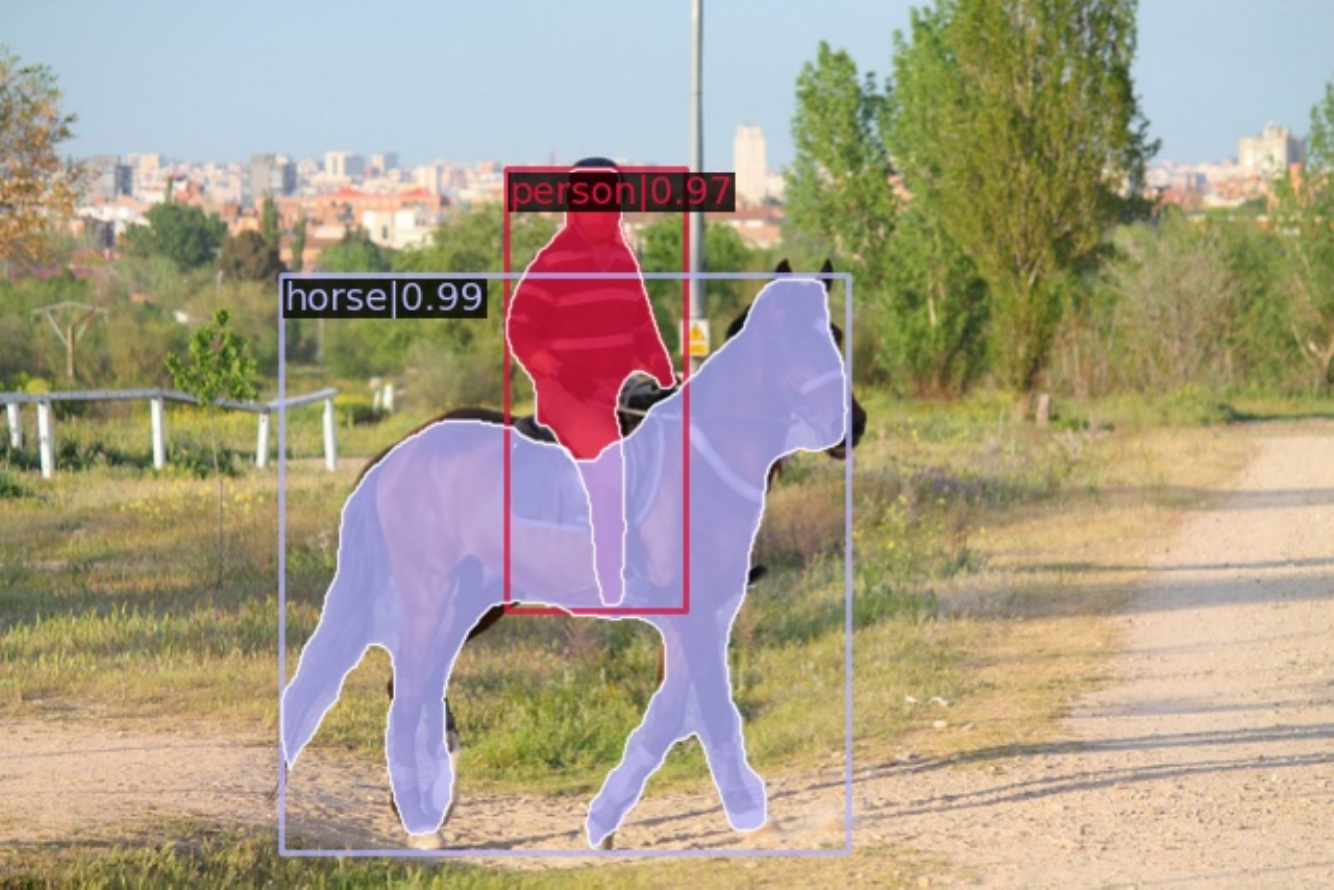}
    \label{ins}
    }
    \quad
    \subfloat[Panoptic Segmentation]{
      \includegraphics[width=1.24in]{fig4-pan.pdf}
      \includegraphics[width=1.24in]{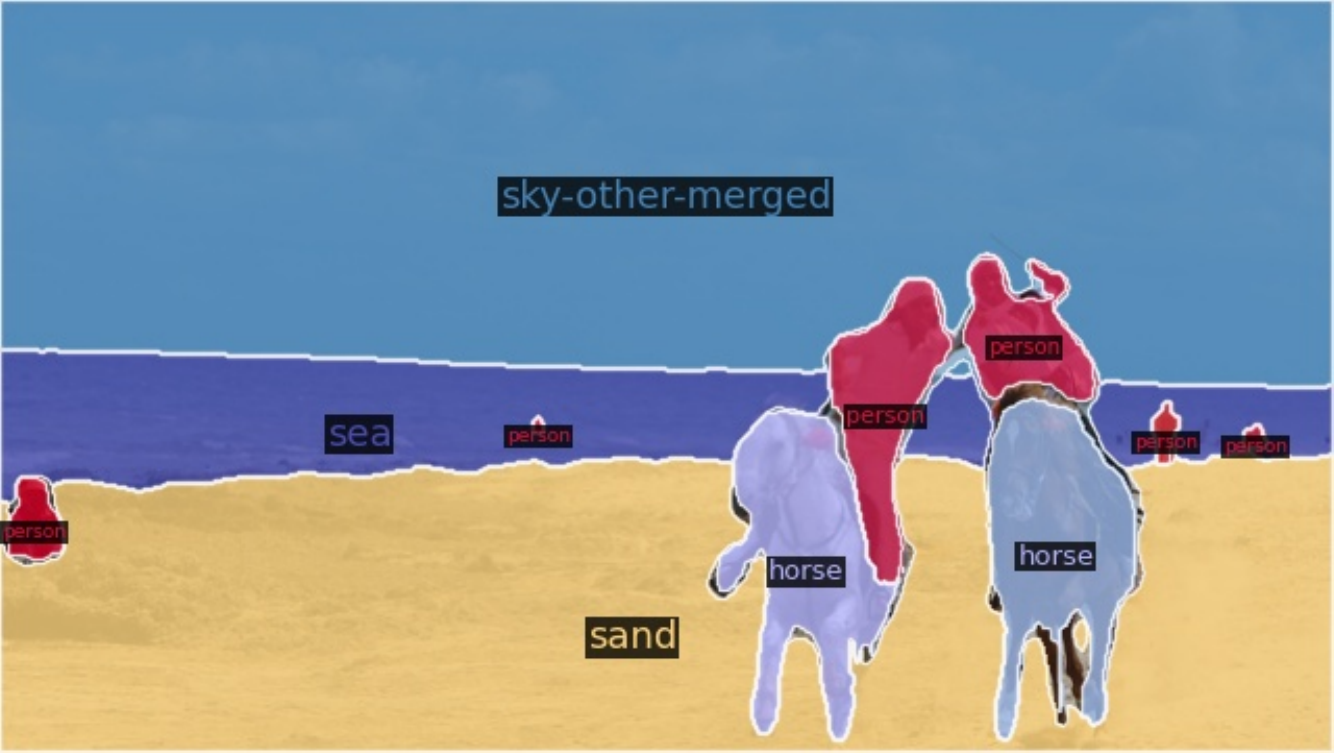}
      \includegraphics[width=1.24in]{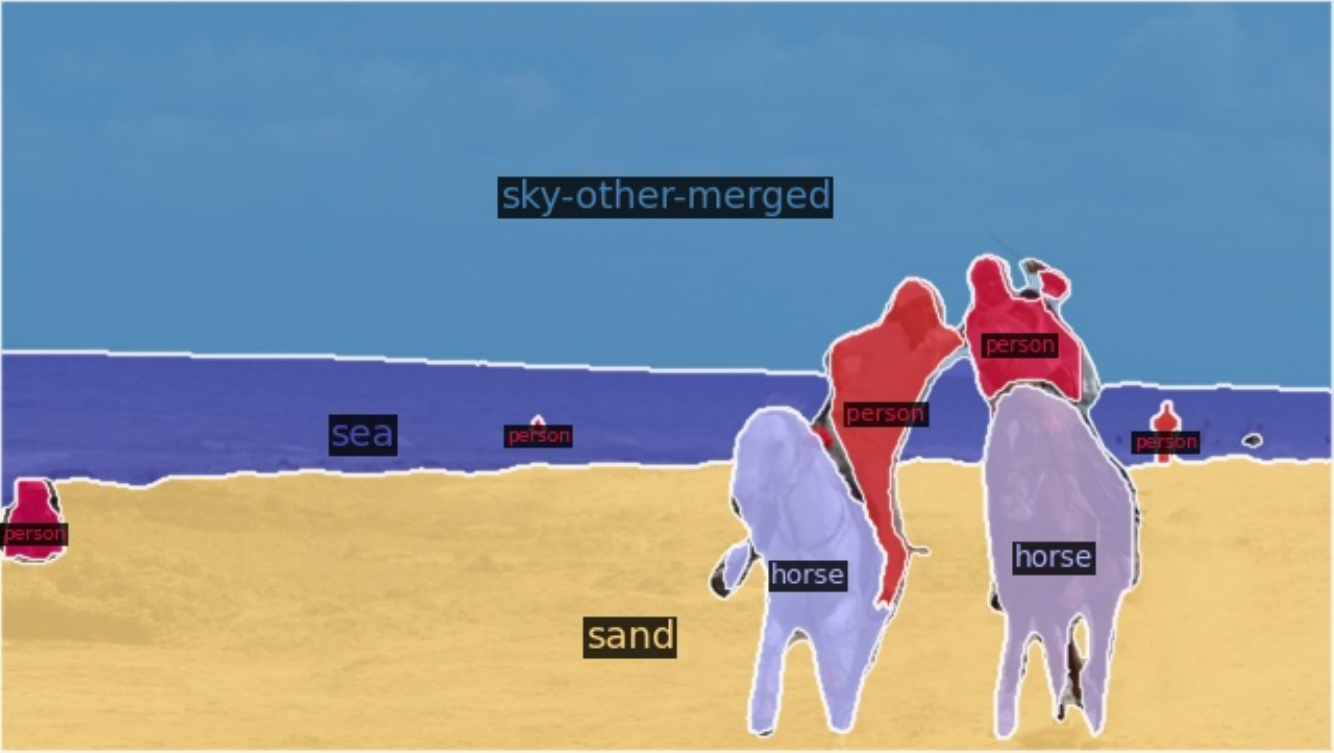}
      \includegraphics[width=1.24in]{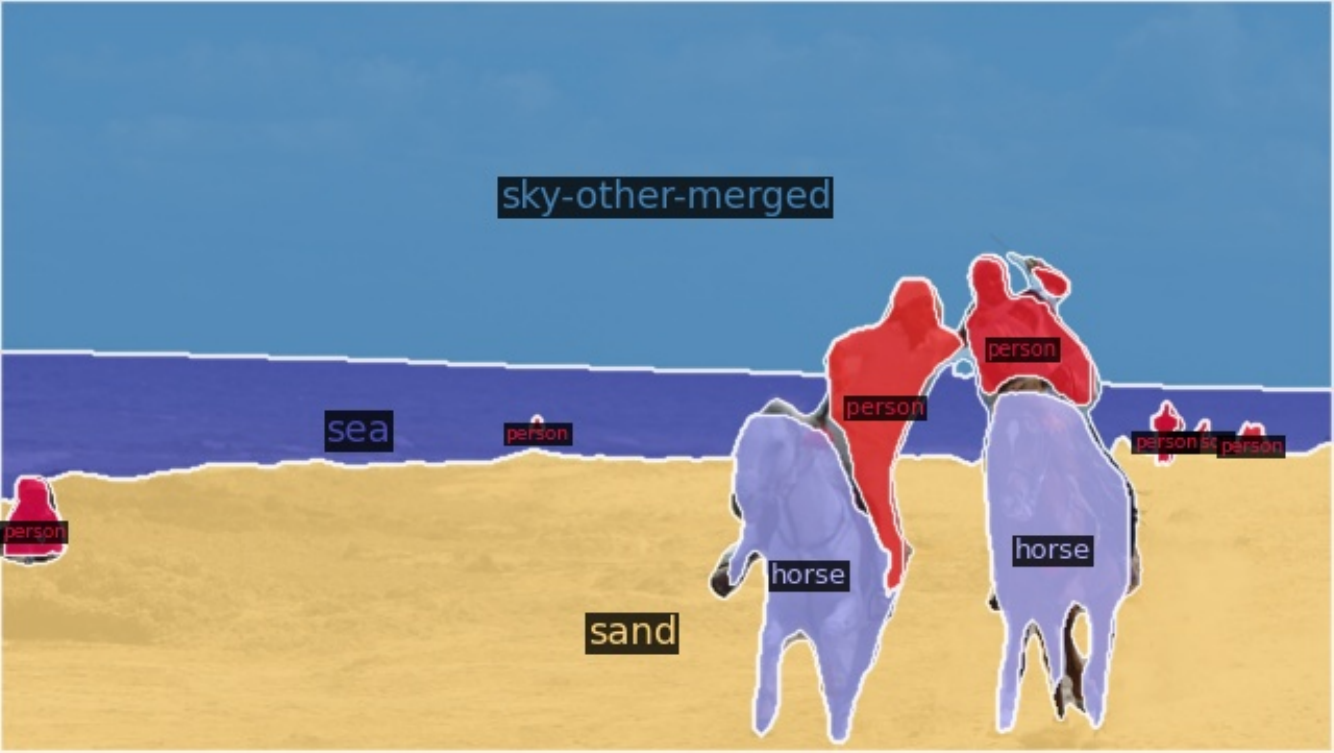}
    \label{pan}
    }
    \quad
    \subfloat[Key-Point Detection]{
      \includegraphics[width=1.24in]{fig4-pose.pdf}
      \includegraphics[width=1.24in]{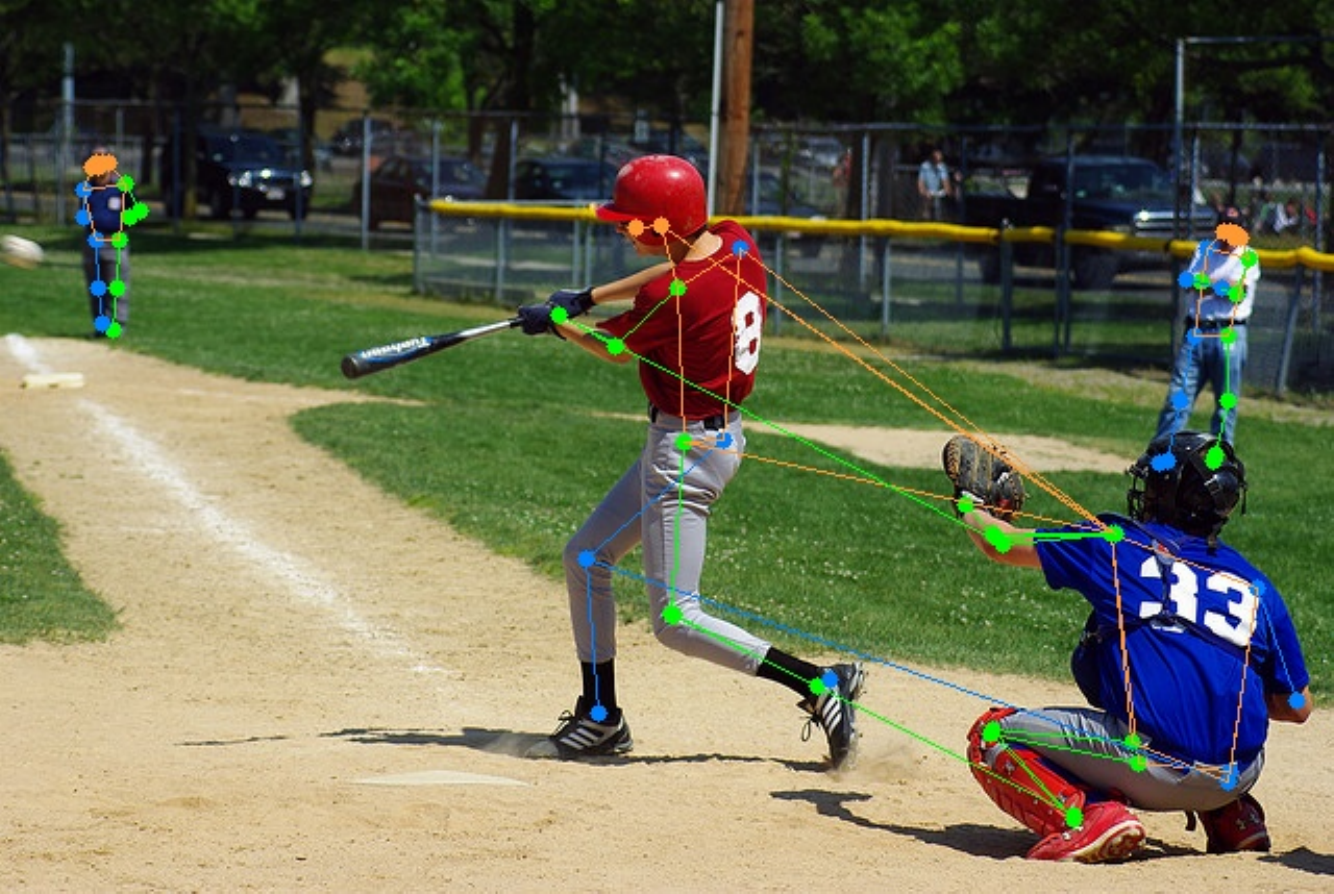}
      \includegraphics[width=1.24in]{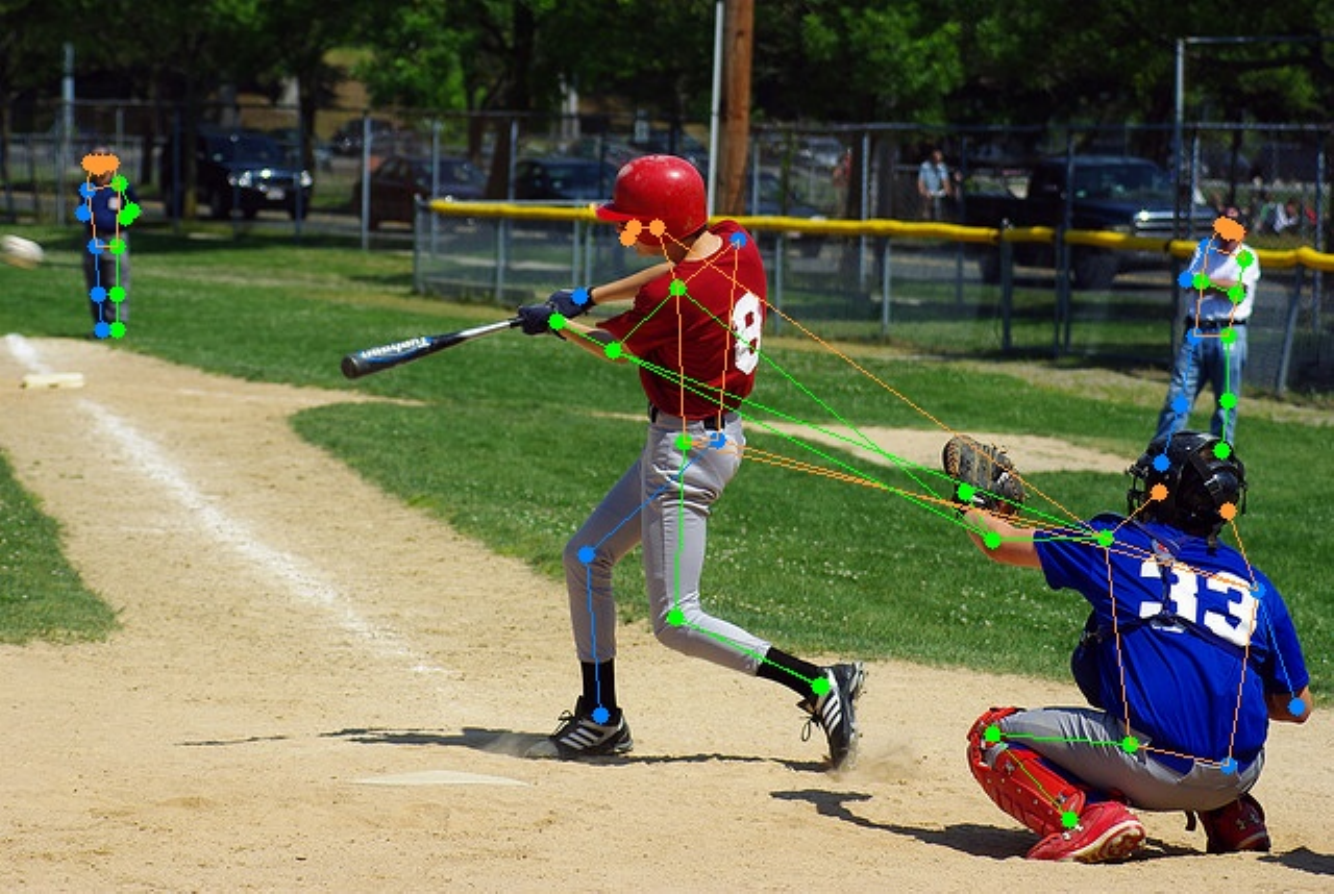}
      \includegraphics[width=1.24in]{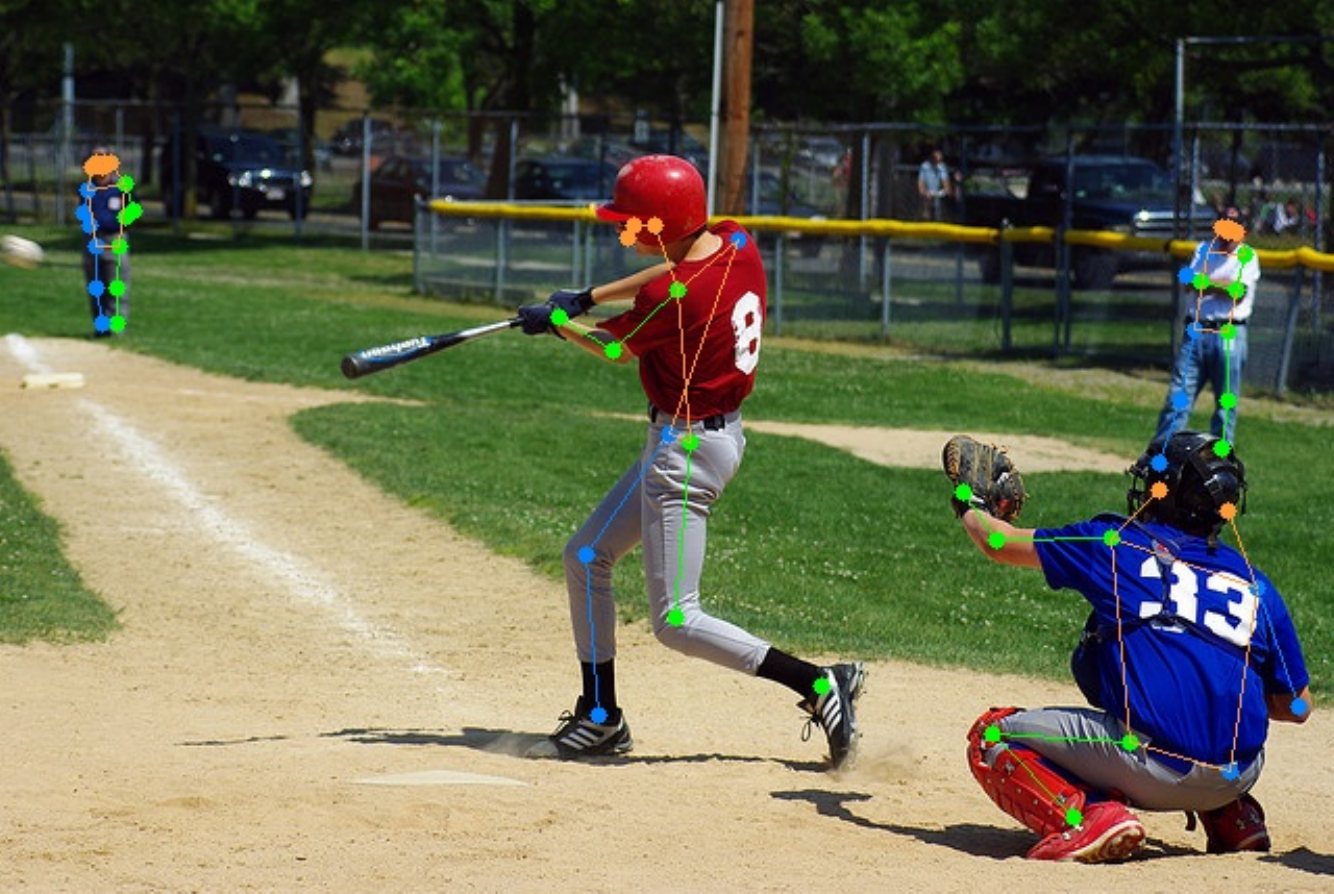}
    \label{key}
    }
    \quad
    \subfloat[Stuff Segmentation]{
      \includegraphics[width=1.24in]{fig4-stuff.pdf}
      \includegraphics[width=1.24in]{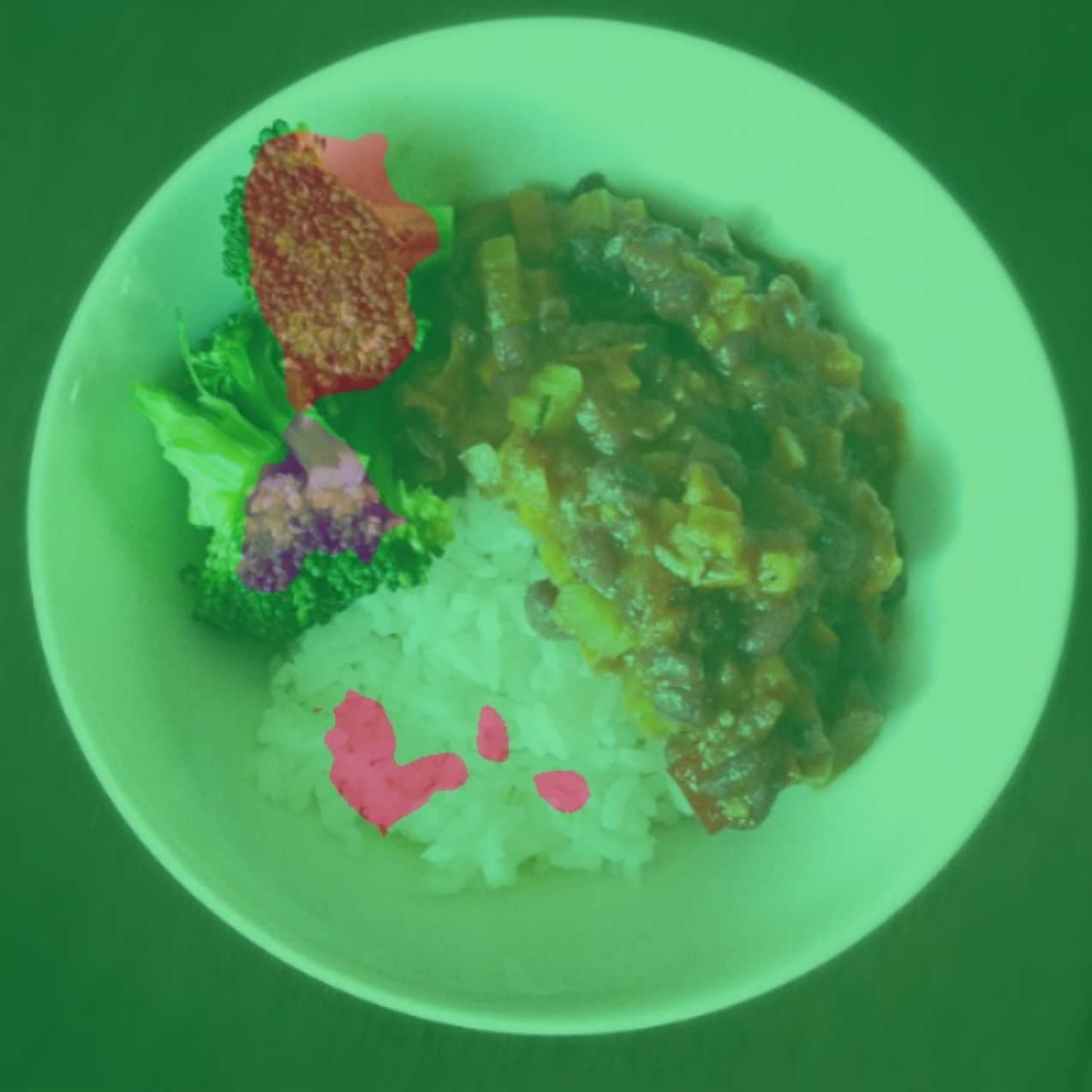}
      \includegraphics[width=1.24in]{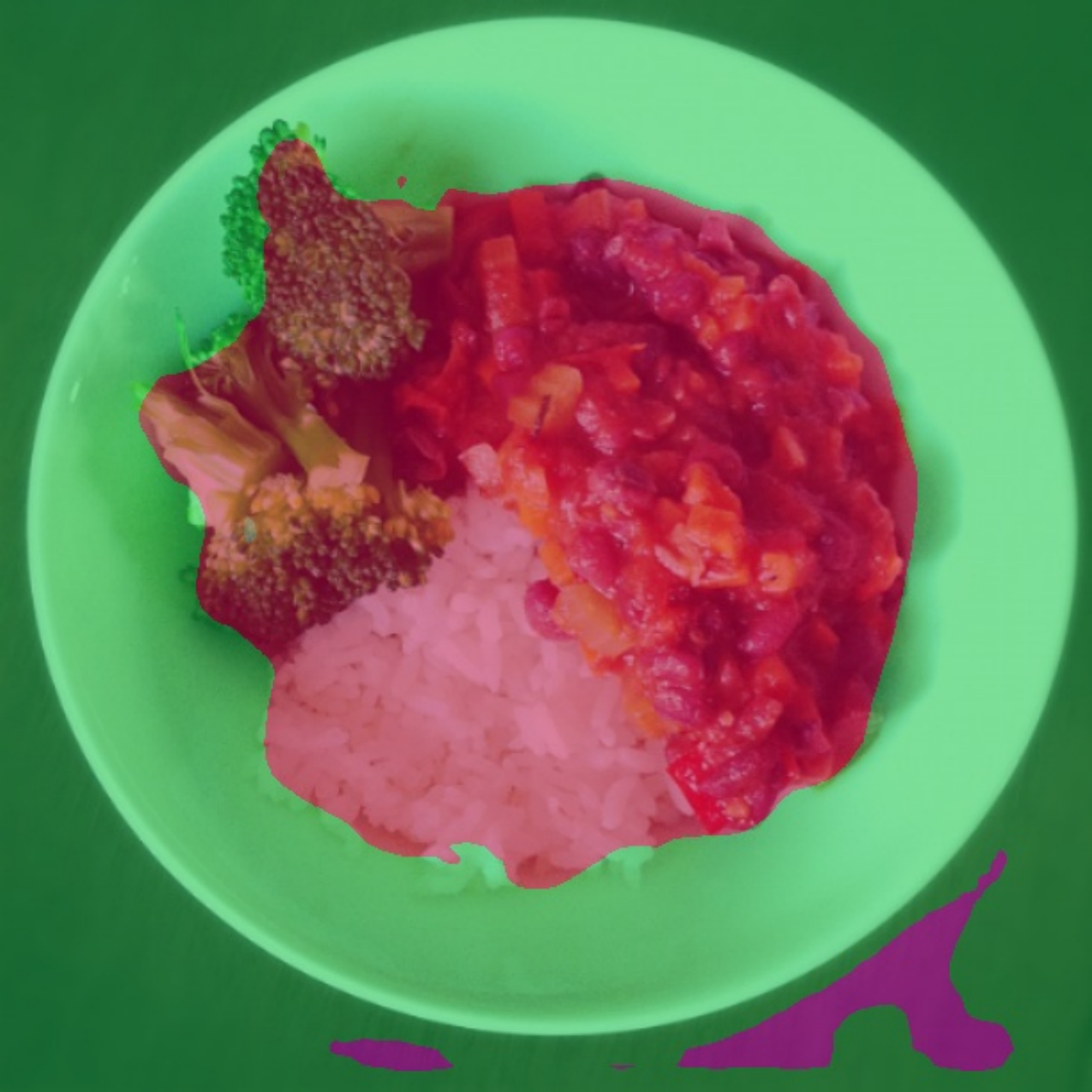}
      \includegraphics[width=1.24in]{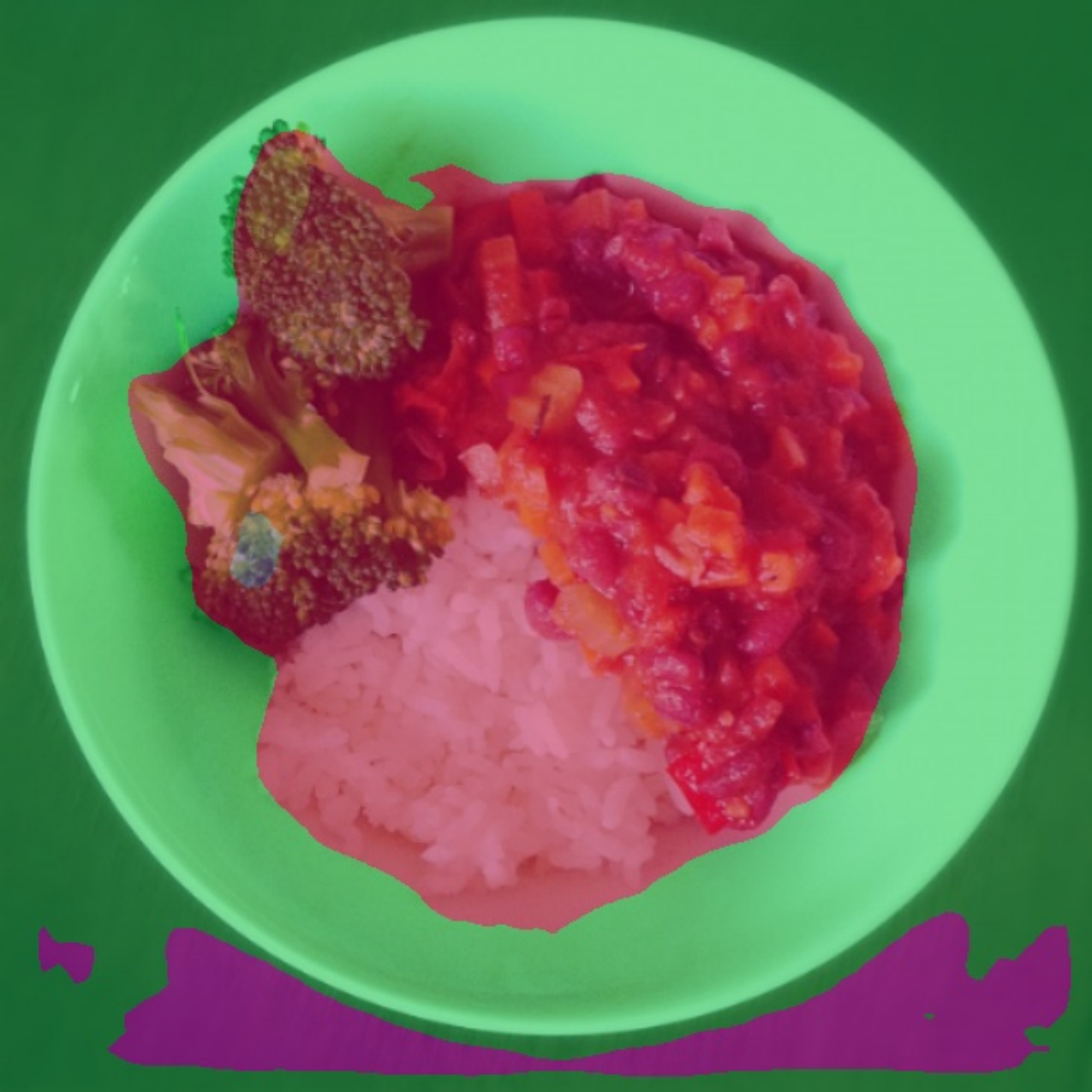}
    \label{stuff}
    }
    \caption{Qualitative results on MS COCO dataset. Each set of results shows the original image, local result, FL result, and our M-Fed result respectively from left to right.}
    \label{quan2}
    \end{figure*}

To further explore the effectiveness of the M-Fed method from a quantitative perspective, we adopt ResNet-50 as the encoder to conduct a quantitative evaluation on the MS COCO dataset and observe its performance. The results are shown in Figure~\ref{quan2}.

As can be seen from Figure~\ref{quan2}, although our method may not achieve the best numerical results across all tasks, it still demonstrates certain advantages in quantitative metrics. Specifically, compared with other methods, our M-Fed method exhibits particularly prominent performance in key-point detection and stuff segmentation tasks. In the key-point detection task, only the M-Fed method can accurately identify the positions of human key-points, while other methods exhibit significant errors during detection. In the stuff segmentation task, our method performs similarly to the FL method but demonstrates finer performance in edge segmentation. In the instance segmentation task, although the score of the M-Fed method is slightly lower than that of the FL method, the result images show that our method is more accurate in distinguishing human contours, especially in recognizing leg contours. Predicting bounding boxes is a relatively new area for other tasks, so the performance of the M-Fed method in unrelated tasks is not outstanding, which warrants further exploration of solutions. In the panoptic segmentation task, both the FL method and our proposed M-Fed method do not perform as well as local methods. However, the M-Fed method can still improve the traditional FL method to some extent. From a quantitative perspective, compared with the FL method, M-Fed demonstrates certain advantages in accurately locating distant human figures.

\section{Discussions}\label{sec6}
The M-Fed method has broken the limitations of traditional FL, which requires uniformity in client models and tasks participating in FL, and has demonstrated performance improvements compared to traditional FL. However, this method still exhibits certain limitations. Specifically: Firstly, although M-Fed allows for the existence of different model structures, clients performing the same task must maintain the same model structure, otherwise intra-task knowledge fusion cannot be achieved; simultaneously, the encoder parts of client models performing different tasks must also maintain the same structure, otherwise inter-task knowledge fusion cannot be realized. Secondly, this method allows clients performing different tasks to participate in collaborative learning, but if the correlation between tasks is too low, it may negatively impact task performance. Although setting thresholds can mitigate this impact, this process requires human involvement and judgment, thereby reducing the autonomy of the method.

\section{Conclusion}\label{sec7}
This paper proposes a novel M-Fed method, aiming to enable multi-task and heterogeneous model clients to participate in FL, breaking the basic assumptions of traditional FL regarding task and model consistency. The core of this method lies in utilizing the widely applicable encoder-decoder structure to achieve multi-task FL. In the M-Fed framework, knowledge fusion among the same tasks is achieved by replacing local models with aggregated task global models. As for knowledge transfer between different tasks, during the client training process, the global encoder model is used as a regularization term to encourage the encoder part of the local model to align with the global encoder model. To mitigate the potential negative impact of the M-Fed method during the early stages of model training, we have designed a variable threshold to regulate its influence. This threshold strategy allows the global encoder model to have minimal involvement in the early stages of training, gradually increasing its participation as the training process progresses. Empirical results and comprehensive analysis have verified the effectiveness of the M-Fed method. Compared with traditional methods, our method not only allows multi-task clients to participate in FL but also demonstrates significant performance advantages.
The M-Fed method represents an innovative attempt to challenge the fundamental assumptions of traditional FL methods. Despite its limitations, it has contributed to the broader application of FL. In future research, we will continue to focus on these limitations and strive to overcome them.

\section*{Acknowledgments}
This work was supported in part by the National Natural Science Foundation of China under Grants 62002369 and 62102445, in part by the Postgraduate Scientific Research Innovation Project of Hunan Province Grant XJCX2023013.

\begin{appendices}

\section{Proof of the Convergence of Algorithm~\ref{alg}}\label{app}
In this section, we conduct an in-depth analysis of the convergence behavior of Algorithm~\ref{alg}. The design of the global encoder model $g$ is independent of any specific task model ${v_{k}}$, where $k \in K$.
Consequently, model $g$ demonstrates a global convergence rate that is consistent with local objectives. Based on this observation, we further derive the local convergence property of Algorithm~\ref{alg}. To comprehensively analyze the convergence characteristics of Algorithm~\ref{alg}, we first outline the following fundamental assumptions:
\begin{itemize}
    \item{For $k\in \left [ K \right ] $, $F_k$ is $\mu $-strongly convex.}
    \item{The expectation of stochastic gradients is uniformly bounded at all devices and all iterations, i.e.,\\ 
  \begin{equation}
    \mathbb{E} \left \| \nabla F(v_k^t) \right \| ^2\le G^2
  \end{equation}
    }
  \end{itemize}

\newtheorem{Lemma}{\bf Lemma}
\begin{Lemma}\label{Lemma1}
Under assumptions above, we have:
\begin{equation}
\begin{aligned}
\mathbb{E}\left \| v_k^{t+1}-v_k^\ast  \right \| ^2 & \le \left ( 1-\eta \mu  \right ) \mathbb{E}\left \| v_k^t-v_k^\ast \right \|^2 \\
& + \eta ^2\left ( \left ( G+\lambda \left ( \frac{G}{\mu }  \right ) +M \right )^2 \right. \\
& \left. + \lambda ^2\mathbb{E}\left \| w^t-w^\ast  \right \|^2  \right ) \\
& \left. - 2\lambda \left ( G+\lambda \left ( \frac{G}{\mu } +M \right )  \right )\mathbb{E}\left \| w^t-w^\ast  \right \|  \right )  \\
& - 2\eta \left ( \frac{G^2}{2\mu }  \right.\\
& + \lambda \mathbb{E}\left \| v_k^t-v_k^\ast  \right \| \left ( \left ( \frac{G}{\mu }+M  \right ) \right.\\
& \left. \left. - \mathbb{E}\left \| w^t-w^\ast  \right \| \right )  \right ) 
\end{aligned}
\end{equation}
\end{Lemma} 

\begin{proof}[Proof of Lemma~\ref{Lemma1}]
    Denote $g(v_k^t;w^t )$ as the stochastic gradient of $h(v_k^t;w^t )$. 
  \begin{equation} \label{fml1}
  \begin{aligned}
    \mathbb{E}\left\| {v_{k}^{t + 1} - v_{k}^{*}} \right\|^{2} & = \mathbb{E}\left\| {v_{k}^{t} - \eta g\left( {v_{k}^{t};w^{t}} \right) - v_{k}^{*}} \right\|^{2} \\
    & = \mathbb{E}\left\| {v_{k}^{t} - v_{k}^{*}} \right\|^{2} + \eta^{2}\underset{A}{\underbrace{\mathbb{E}\left\| {g\left( {v_{k}^{t};w^{t}} \right)} \right\|^{2}}} \\
    & - 2\eta\underset{B}{\underbrace{\mathbb{E}\left\langle {v_{k}^{t} - v_{k}^{*},g\left( {v_{k}^{t};w^{t}} \right)} \right\rangle}}
  \end{aligned}
  \end{equation}
  
  Among $A$,
  \begin{equation}
  \begin{aligned}
  A & = \mathbb{E}\left\| {g\left( {v_{k}^{t};w^{t}} \right)} \right\|^{2} = \mathbb{E}\left\| {\nabla F\left( v_{k}^{t} \right) + \lambda\left( v_{k}^{g,t} - w^{t} \right)} \right\|^{2} \\
  & = \mathbb{E}\left\| {\nabla F\left( v_{k}^{t} \right) + \lambda v_{k}^{g,t} - {\lambda w}^{t} - {\lambda w}^{*} + {\lambda w}^{*}} \right\|^{2} \\
  & = \mathbb{E}\left\| {\nabla F\left( v_{k}^{t} \right) + \lambda\left( v_{k}^{g,t} - w^{*} \right)\left. + {\lambda\left( w \right.}^{*} - w^{t} \right)} \right\|^{2} \\
  & = \mathbb{E}\left\| \left. g\left( {v_{k}^{t};w^{*}} \right) - {\lambda\left( w \right.}^{t} - w^{*} \right) \right\|^{2} \\
  & = \underset{A^{1}}{\underbrace{\mathbb{E}\left\| {g\left( {v_{k}^{t};w^{*}} \right)} \right\|^{2}}} + \lambda^{2}\mathbb{E}\left\| {w^{t} - w^{*}} \right\|^{2} \\
  & - 2\lambda\mathbb{E}\left\langle {g\left( {v_{k}^{t};w^{*}} \right),w^{t} - w^{*}} \right\rangle
  \end{aligned}
  \end{equation}
  
  Among $A^{1}$,
  \begin{equation}
  \begin{aligned}
  A^{1} & = \mathbb{E}\left\| {g\left( {v_{k}^{t};w^{*}} \right)} \right\|^{2} = \mathbb{E}\left\| {\nabla F\left( v_{k}^{t} \right) + \lambda\left( v_{k}^{g,t} - w^{*} \right)} \right\|^{2}\\
  & = \mathbb{E}\left\| {\nabla F\left( v_{k}^{t} \right)} \right\|^{2} + \lambda^{2}\mathbb{E}\left\| {v_{k}^{g,t} - w^{*}} \right\|^{2} \\
  & + 2\lambda\mathbb{E}\left\langle {\nabla F\left( v_{k}^{t} \right),v_{k}^{g,t} - w^{*}} \right\rangle
  \end{aligned}
  \end{equation}
  
  Further, note that,
  \begin{equation}
  \mathbb{E}\left\| {\nabla F\left( v_{k}^{t} \right)} \right\|^{2} \leq G^{2}
  \end{equation}
  \begin{equation}
  \begin{aligned}
  \mathbb{E}\left\| {v_{k}^{g,t} - w^{*}} \right\|^{2} & = \mathbb{E}\left\| {v_{k}^{g,t} - u_{k}^{*} + u_{k}^{*} - w^{*}} \right\|^{2}\\
  & = \mathbb{E}\left\| {v_{k}^{g,t} - u_{k}^{*}} \right\|^{2} + \mathbb{E}\left\| {u_{k}^{*} - w^{*}} \right\|^{2} \\
  & + 2\mathbb{E}\left\langle {v_{k}^{g,t} - u_{k}^{*},u_{k}^{*} - w^{*}} \right\rangle
  \end{aligned}
  \end{equation}
  
  According to the property of $\mu $-strongly convexity, we have:
  \begin{gather}
  \mathbb{E}\left\| {v_{k}^{g,t} - u_{k}^{*}} \right\|^{2} \leq \frac{G^{2}}{\mu^{2}}\\
  \mathbb{E}\left\| {u_{k}^{*} - w^{*}} \right\|^{2} \leq M^{2}\\
  \mathbb{E}\left\langle {v_{k}^{g,t} - u_{k}^{*},u_{k}^{*} - w^{*}} \right\rangle \leq \frac{G}{\mu}M
  \end{gather}
  
  Therefore,
  \begin{equation}
  \mathbb{E}\left\| {v_{k}^{g,t} - w^{*}} \right\|^{2} \leq \left( \frac{G}{\mu} + M \right)^{2}
  \end{equation}
  \begin{equation}
  \begin{aligned}
  A^{1} = \mathbb{E}\left\| {g\left( {v_{k}^{t};w^{*}} \right)} \right\|^{2} \leq G^{2} + \lambda^{2}\left( {\frac{G}{\mu} + M} \right)^{2} + 2\lambda G\left( \frac{G}{\mu} + M \right)
  \end{aligned}
  \end{equation}
  \begin{equation}
  \begin{aligned}
  A & \leq \left( {G + \lambda\left( {\frac{G}{\mu} + M} \right)} \right)^{2} + \lambda^{2}\mathbb{E}\left\| {w^{t} - w^{*}} \right\|^{2} \\
  & - 2\lambda\left( {G + \lambda\left( {\frac{G}{\mu} + M} \right)} \right)\mathbb{E}\left\| {w^{t} - w^{*}} \right\|
  \end{aligned}
  \end{equation}
  
  Next consider $B$:
  \begin{gather}
  B = \mathbb{E}\left\langle {v_{k}^{t} - v_{k}^{*},g\left( {v_{k}^{t};w^{t}} \right)} \right\rangle\\
  = \mathbb{E}\left\langle {v_{k}^{t} - v_{k}^{*},\nabla F\left( v_{k}^{t} \right) + \lambda\left( v_{k}^{g,t} - w^{t} \right)} \right\rangle\\
  = \underset{B^{1}}{\underbrace{\mathbb{E}\left\langle {v_{k}^{t} - v_{k}^{*},\nabla F\left( v_{k}^{t} \right)} \right\rangle}} + \lambda\underset{B^{2}}{\underbrace{\mathbb{E}\left\langle {v_{k}^{t} - v_{k}^{*},v_{k}^{g,t} - w^{t}} \right\rangle}}
  \end{gather}
  
  According to the property of $\mu $-strongly convexity:
  \begin{equation}
  \begin{aligned}
  B^{1} & = \mathbb{E}\left\langle {v_{k}^{t} - v_{k}^{*},\nabla F\left( v_{k}^{t} \right)} \right\rangle \\
  & \leq \mathbb{E}\left( F\left( v_{k}^{t} \right) - F\left( v_{k}^{*} \right) + \frac{\mu}{2}\left\| {v_{k}^{t} - v_{k}^{*}} \right\|^{2} \right)
  \end{aligned}
  \end{equation}
  
  According to \emph{Polyak Lojasiewicz (PL)} condition:
  \begin{equation}
  F\left( v_{k}^{t} \right) - F\left( v_{k}^{*} \right) \leq \frac{1}{2\mu}\left\| {\nabla F\left( v_{k}^{t} \right)} \right\|^{2}
  \end{equation}
  
  Therefore,
  \begin{equation}
  \begin{aligned}
  B^{1} & \le \frac{1}{2\mu } \mathbb{E} \left \| \nabla F\left ( v_{k}^{t}   \right )  \right \| ^{2}+\frac{\mu }{2}\mathbb{E} \left \| v_{k}^{t}-v_{k}^{\ast }  \right \|^{2} \\
  & \leq \frac{G^{2}}{2\mu} + \frac{\mu}{2}{\mathbb{E}\left\| {v_{k}^{t} - v_{k}^{*}} \right\|}^{2}
  \end{aligned}
  \end{equation}
  
  Considering $B^2$, we have:
  \begin{equation}
  \begin{aligned}
  B^{2} & = \mathbb{E}\left\langle {v_{k}^{t} - v_{k}^{*},v_{k}^{g,t} - w^{t}} \right\rangle \\
  & \leq \mathbb{E}\left\lbrack \left\| {v_{k}^{t} - v_{k}^{*}} \right\|\left\| {v_{k}^{g,t} - w^{t}} \right\| \right\rbrack
  \end{aligned}
  \end{equation}
  
  According to $\mathbb{E}\lbrack XY\rbrack \leq \sqrt{\mathbb{E}\left\lbrack X^{2} \right\rbrack\mathbb{E}\left\lbrack Y^{2} \right\rbrack}$,
  \begin{equation}
  B^{2} \leq \sqrt{\mathbb{E}\left\| {v_{k}^{t} - v_{k}^{*}} \right\|^{2}\mathbb{E}\left\| {v_{k}^{g,t} - w^{t}} \right\|^{2}}
  \end{equation}
  
  Among them,
  \begin{gather}
  \mathbb{E}\left \| v_{k}^{g,t}-w^{t} \right \|^{2}=\mathbb{E}\left \| v_{k}^{g,t}-w^{\ast } +w^{\ast }-w^{t} \right \|^{2}\\
  \begin{matrix}
    {= \mathbb{E}\left\| {v_{k}^{g,t} - w^{*}} \right\|^{2} + \mathbb{E}\left\| {w^{t} - w^{*}} \right\|^{2}} \\
    {- 2\mathbb{E}\left\langle {v_{k}^{g,t} - w^{*},w^{t} - w^{*}} \right\rangle} \\
  \end{matrix} \\
  \begin{matrix}
    {\leq \left( {\frac{G}{\mu} + M} \right)^{2} + \mathbb{E}\left\| {w^{t} - w^{*}} \right\|^{2}} \\
    {- 2\left( {\frac{G}{\mu} + M} \right)\mathbb{E}\left\| {w^{t} - w^{*}} \right\|} \\
  \end{matrix}
  \end{gather}
  
  Therefore,
  \begin{equation}
  \begin{aligned}
    B^{2} & \leq \sqrt{\mathbb{E}\left\| {v_{k}^{t} - v_{k}^{*}} \right\|^{2}\left( B^{3} - B^{4} \right)} \\
    B^{3} & = \left( {\frac{G}{\mu} + M} \right)^{2} + \mathbb{E}\left\| {w^{t} - w^{*}} \right\|^{2} \\
    B^{4} & = 2\left( {\frac{G}{\mu} + M} \right)\mathbb{E}\left\| {w^{t} - w^{*}} \right\|
  \end{aligned}
  \end{equation}
  
  Thus the bound of $B$ is obtained:
  \begin{equation}
  B \leq \frac{G^{2}}{2\mu} + \frac{\mu}{2}{\mathbb{E}\left\| {v_{k}^{t} - v_{k}^{*}} \right\|}^{2} \lambda\sqrt{\mathbb{E}\left\| {v_{k}^{t} - v_{k}^{*}} \right\|^{2} \left( B^{3} - B^{4} \right)}
  \end{equation}
  
  Plug it into formula~\ref{fml1},
  \begin{equation}
    \begin{aligned}
      \mathbb{E}\left\| {v_{k}^{t + 1} - v_{k}^{*}} \right\|^{2} & \leq \mathbb{E}\left\| {v_{k}^{t} - v_{k}^{*}} \right\|^{2} + \eta^{2}\left( A^{2} - A^{3} \right) \\
      & - 2\eta\left( \frac{G^{2}}{2\mu} + \frac{\mu}{2}{\mathbb{E}\left\| {v_{k}^{t} - v_{k}^{*}} \right\|}^{2} \right. \\
      & \left. + \lambda\sqrt{\mathbb{E}\left\| {v_{k}^{t} - v_{k}^{*}} \right\|^{2}\left( B^{3} - B^{4} \right)} \right) \\
    \end{aligned}
    \end{equation}
  
    \begin{equation}
      \begin{aligned}
    A^{2} & = \left( {G + \lambda\left( {\frac{G}{\mu} + M} \right)} \right)^{2} + \lambda^{2}\mathbb{E}\left\| {w^{t} - w^{*}} \right\|^{2} \\
    A^{3} & = 2\lambda\left( {G + \lambda\left( {\frac{G}{\mu} + M} \right)} \right)\mathbb{E}\left\| {w^{t} - w^{*}} \right\|
  \end{aligned}
  \end{equation}
  
  Among them,
  \begin{equation}
  \begin{aligned}
  & = (1 - \eta\mu)\mathbb{E}\left\| {v_{k}^{t} - v_{k}^{*}} \right\|^{2} + \eta^{2}\left( \left( {G + \lambda\left( {\frac{G}{\mu} + M} \right)} \right)^{2} \right. \\
  & + \lambda^{2}\mathbb{E}\left\| {w^{t} - w^{*}} \right\|^{2} \\
  & \left. - 2\lambda\left( {G + \lambda\left( {\frac{G}{\mu} + M} \right)} \right)\mathbb{E}\left\| {w^{t} - w^{*}} \right\| \right) \\
  & - 2\eta\left( \frac{G^{2}}{2\mu} \right. \\
  & \left. + \lambda\mathbb{E}\left\| {v_{k}^{t} - v_{k}^{*}} \right\|\left( {\left( {\frac{G}{\mu} + M} \right) - \mathbb{E}\left\| {w^{t} - w^{*}} \right\|} \right) \right)
  \end{aligned}
  \end{equation}
  
  Lemma~\ref{Lemma1} is proved.
  \end{proof}

\begin{theorem}\label{theorem}
Under the assumptions above, there is a constant $A$, so that $\frac{g(t + 1)}{g(t)} \geq 1 - \frac{g(t)}{A}$, there is a constant $C \leq \infty$, for all clients $k \in \lbrack K\rbrack$, there is $\mathbb{E}\left\lbrack \left\| {v_{k}^{t} - v_{k}^{*}} \right\|^{2} \right\rbrack \leq Cg(t)$, where the learning rate $\eta = \frac{2g(t)}{A\mu}$.
\end{theorem} 

\begin{proof}[Proof of Theorem~\ref{theorem}]
    We prove it by induction. When $ t = 0 $, for any constant $C > \frac{\mathbb{E}\left\| {v_{k}^{0} - v_{k}^{*}} \right\|^{2}}{g(0)}$, we have $\mathbb{E}\left\| {v_{k}^{0} - v_{k}^{*}} \right\|^{2} \leq Cg(0)$. When $\mathbb{E}\left\| {v_{k}^{t} - v_{k}^{*}} \right\|^{2} \leq Cg(t)$ is true, for $t + 1$, according to Lemma~\ref{Lemma1}, there are:
    \begin{equation}
    \begin{aligned}
    \mathbb{E}\left\| {v_{k}^{t + 1} - v_{k}^{*}} \right\|^{2} & \leq \left( {1 - \frac{2g(t)}{A}} \right)Cg(t) \\
    & + \frac{g^{2}(t)}{A}\frac{4}{A\mu^{2}}\left( \left( {G + \lambda\left( {\frac{G}{\mu} + M} \right)} \right)^{2} \right. \\
    & \left. + \lambda^{2}g(t) - 2\lambda\left( {G + \lambda\left( {\frac{G}{\mu} + M} \right)} \right)\sqrt{g(t)} \right) \\
    & - \frac{4g(t)}{A\mu}\left( \frac{G^{2}}{2\mu} \right. \\
    & \left. + \lambda\sqrt{Cg(t)}\left( {\left( {\frac{G}{\mu} + M} \right) - \sqrt{g(t)}} \right) \right)
    \end{aligned}
    \end{equation}
    
    \begin{equation}
    \begin{aligned}
    & = \left( {1 - \frac{2g(t)}{A}} \right)Cg(t)\\
    & + \frac{g^{2}(t)}{A}\frac{4}{A\mu^{2}}\begin{pmatrix}
      {\left( {G + \lambda\left( {\frac{G}{\mu} + M} \right)} \right)^{2} + \lambda^{2}g(t)} \\
      {- 2\lambda\left( {G + \lambda\left( {\frac{G}{\mu} + M} \right)} \right)\sqrt{g(t)}} \\
      \end{pmatrix} \\
    & - \frac{4g(t)}{A\mu}\left( \frac{G^{2}}{2\mu} + \lambda\sqrt{Cg(t)}\left( {\frac{G}{\mu} + M} \right) - \lambda\sqrt{C}g(t) \right)
    \end{aligned}
    \end{equation}
    
    \begin{equation}
    \begin{aligned}
    & = \left( {1 - \frac{2g(t)}{A}} \right)Cg(t) \\
    & + \frac{g^{2}(t)}{A}\frac{4}{A\mu^{2}}\begin{pmatrix}
      {\left( {G + \lambda\left( {\frac{G}{\mu} + M} \right)} \right)^{2} + \lambda^{2}g(t)} \\
      {- 2\lambda\left( {G + \lambda\left( {\frac{G}{\mu} + M} \right)} \right)\sqrt{g(t)}} \\
      \end{pmatrix} \\
    & - \frac{g^{2}(t)}{A}\left( \frac{{4G}^{2}}{2\mu^{2}g(t)} + \frac{4\lambda\sqrt{C}\left( {\frac{G}{\mu} + M} \right)}{\mu\sqrt{g(t)}} - \frac{4\lambda\sqrt{C}}{\mu} \right)
    \end{aligned}
    \end{equation}
    
    \begin{equation}
    \begin{aligned}
    & \leq \left( {1 - \frac{2g(t)}{A}} \right)Cg(t) \\
    & + \frac{g^{2}(t)}{A}\left( \frac{\begin{matrix}
      {4\left( {G + \lambda\left( {\frac{G}{\mu} + M} \right)} \right)^{2}} \\
      {+ 4\lambda^{2}g(t) - 8\lambda\left( {G + \lambda\left( {\frac{G}{\mu} + M} \right)} \right)\sqrt{g(t)}} \\
      \end{matrix}}{A\mu^{2}} \right.\\
    & \left. - \frac{{4G}^{2}}{2\mu^{2}g(t)} + \frac{4\lambda\sqrt{C}}{\mu}\left( 1 - \frac{\left( {\frac{G}{\mu} + M} \right)}{\sqrt{g(t)}} \right) \right)
    \end{aligned}
    \end{equation}
    
    \begin{equation}
    \begin{aligned}
    & \leq \left( {1 - \frac{2g(t)}{A}} \right)Cg(t) + C\frac{g^{2}(t)}{A}\\
    & = \left( {1 - \frac{2g(t)}{A} + \frac{g(t)}{A}} \right)Cg(t)\\
    & = \left( {1 - \frac{g(t)}{A}} \right)Cg(t)
    \end{aligned}
    \end{equation}
    
    Because $\frac{g(t + 1)}{g(t)} \geq 1 - \frac{g(t)}{A}$ :
    \begin{equation}
    \begin{aligned}
    & \leq \frac{g\left( {t + 1} \right)}{g(t)}Cg(t)\\
    & = Cg\left( {t + 1} \right)
    \end{aligned}
    \end{equation}
    
    completing the proof.
    \end{proof}

\end{appendices}


\bibliography{reference.bib}

\end{document}